\documentclass[sigconf]{acmart}
\AtBeginDocument{%
  }

\usepackage{subfigure}
\usepackage{multirow}
\usepackage{multicol}
\usepackage[linesnumbered,boxed,ruled,commentsnumbered]{algorithm2e}
\usepackage{enumitem}
\setlist[itemize]{leftmargin=0.4cm}
\usepackage{pifont}
\newcommand{\para}[1]{\vspace{2mm} \noindent \textbf{#1}}
\usepackage{xcolor}
\usepackage{color, colortbl}
\definecolor{yymgray1}{HTML}{DDDFE8}
\definecolor{yymgray2}{HTML}{F2F2F2}
\definecolor{yympurple}{HTML}{AA93B4}
\definecolor{yymblue}{HTML}{E5F2FC}
\definecolor{yymblue1}{rgb}{0.6, 0.85, 0.95}
\definecolor{yymorange}{HTML}{FDECEE}
\definecolor{yymgreen}{HTML}{e6faef}
\definecolor{yymgreen1}{rgb}{0.7, 0.85, 0.85}

\begin{document}

\title{Beyond Single-Granularity Prompts: A Multi-Scale Chain-of-Thought Prompt Learning for Graph}


\author{Ziyu Zheng}
\orcid{https://orcid.org/0009-0000-3662-0832}
\affiliation{%
  \department{School of Computer Science and Technology,}
  \institution{Xidian University,}
  \city{Xi'an}
  \country{China}
}
\email{zhengziyu@stu.xidian.edu.cn}

\author{Yaming Yang}
\orcid{https://orcid.org/0000-0002-8186-0648}
\affiliation{%
  \department{School of Computer Science and Technology,}
  \institution{Xidian University,}
  \city{Xi'an}
  \country{China}
}
\email{yym@xidian.edu.cn}

\author{Ziyu Guan}
\orcid{https://orcid.org/0000-0003-2413-4698}
\affiliation{%
  \department{School of Computer Science and Technology,}
  \institution{Xidian University,}
  \city{Xi'an}
  \country{China}
}
\email{zyguan@xidian.edu.cn}

\author{Wei Zhao}
\authornote{Corresponding Author}
\orcid{https://orcid.org/0000-0002-9767-1323}
\affiliation{%
  \department{School of Computer Science and Technology,}
  \institution{Xidian University,}
  \city{Xi'an}
  \country{China}
}
\email{ywzhao@mail.xidian.edu.cn}

\author{Xinyan Huang}
\orcid{https://orcid.org/0009-0002-2057-7432}
\affiliation{%
  \department{School of Artificial Intelligence,}
  \institution{Xidian University,}
  \city{Xi'an}
  \country{China}
}
\email{	xinyanh@stu.xidian.edu.cn}

\author{Weigang Lu}
\orcid{https://orcid.org/0000-0003-4855-7070}
\affiliation{%
  \department{Department of Civil and Environmental Engineering,}
  \institution{The Hong Kong University of Science and Technology,}
  \country{Hong Kong SAR}
}
\email{weiganglu314@outlook.com}
 \renewcommand{\shortauthors}{Ziyu Zheng, et al.}

\begin{abstract}
The ``pre-train, prompt" paradigm, designed to bridge the gap between pre-training tasks and downstream objectives, has been extended from the NLP domain to the graph domain and has achieved remarkable progress. Current mainstream graph prompt-tuning methods modify input or output features using learnable prompt vectors. However, existing approaches are confined to single-granularity (e.g., node-level or subgraph-level) during prompt generation, overlooking the inherently multi-scale structural information in graph data, which limits the diversity of prompt semantics. To address this issue, we pioneer the integration of multi-scale information into graph prompt and propose a Multi-Scale Graph Chain-of-Thought (MSGCOT) prompting framework. Specifically, we design a lightweight, low-rank coarsening network to efficiently capture multi-scale structural features as hierarchical basis vectors for prompt generation. Subsequently, mimicking human cognition from coarse-to-fine granularity, we dynamically integrate multi-scale information at each reasoning step, forming a progressive coarse-to-fine prompt chain. Extensive experiments on eight benchmark datasets demonstrate that MSGCOT outperforms the state-of-the-art single-granularity graph prompt-tuning method, particularly in few-shot scenarios, showcasing superior performance. The code is available at: \url{https://github.com/zhengziyu77/MSGCOT}.
\end{abstract}

\ccsdesc[500]{Information systems~Web mining}
\ccsdesc[500]{Computing methodologies~Unsupervised learning}
\keywords{Graph Neural Networks; Graph Prompt Learning; Few-Shot Learning}

\maketitle

\section{Introduction}
Graph Neural Networks (GNNs) have been widely adopted in real-world scenarios, such as social networks~\cite{social}, anomaly detection~\cite{anomaly_dete}, and recommendation systems~\cite{graph_recom}, due to their ability to capture complex structural dependencies among data. In recent years, the scarcity of labels in practical settings has spurred extensive research on the ``pre-training and fine-tuning" paradigm for GNNs~\cite{pretrain,graphcl,discrepancyMAE,shgp}. This paradigm utilizes self-supervised learning to derive generic, task-agnostic representations from unlabeled graphs, followed by fine-tuning the pre-trained model on downstream tasks using task-specific labels. However, these approaches suffer from an inherent limitation: the discrepancy between pre-training objectives and downstream tasks, which leads to suboptimal performance~\cite{GPL_study}.

To address this issue, recent work introduces prompt tuning as an alternative to fine-tuning. Prompt tuning is initially proposed in NLP and achieving remarkable success~\cite{nlpprompt}, and can effectively bridge the gap between pre-training objectives and downstream tasks by incorporating lightweight learnable prompt parameters to modify inputs or outputs, without updating the parameters of the pre-trained model. Inspired by this, pre-training and prompt tuning have been extended to the graph domain~\cite{allinone,GPF,gppt}. Compared to fine-tuning, graph prompt exhibits lower computational overhead and better generalization. 

Existing graph prompt learning methods can be categorized into two categories based on dependency on pre-training strategies. Pre-training-dependent methods~\cite{allinone, graphprompt,dagprompt} design unified templates for pre-training objectives and downstream tasks to reduce their discrepancy. Pre-training-agnostic methods~\cite{GPF,SUPT,gcot,edgeprompt} enhance generality by inserting learnable feature prompt vectors into inputs or outputs, requiring no specific pre-training strategy. Further, we classify pre-training-agnostic methods by their prompting mechanisms: (1) Single-step prompt tuning~\cite{GPF,edgeprompt,SUPT,pronog} employ a direct generation approach where the complete prompt for downstream tasks is produced in a single forward pass. (2) Multi-step prompt tuning~\cite{gcot} progressively derives final prompts through iterative inference, mimicking the Chain-of-Thought (CoT)~\cite{cot} rationale to achieve stronger expressiveness.

\begin{figure*}[h]
\centering   
\includegraphics[width=0.98\linewidth]{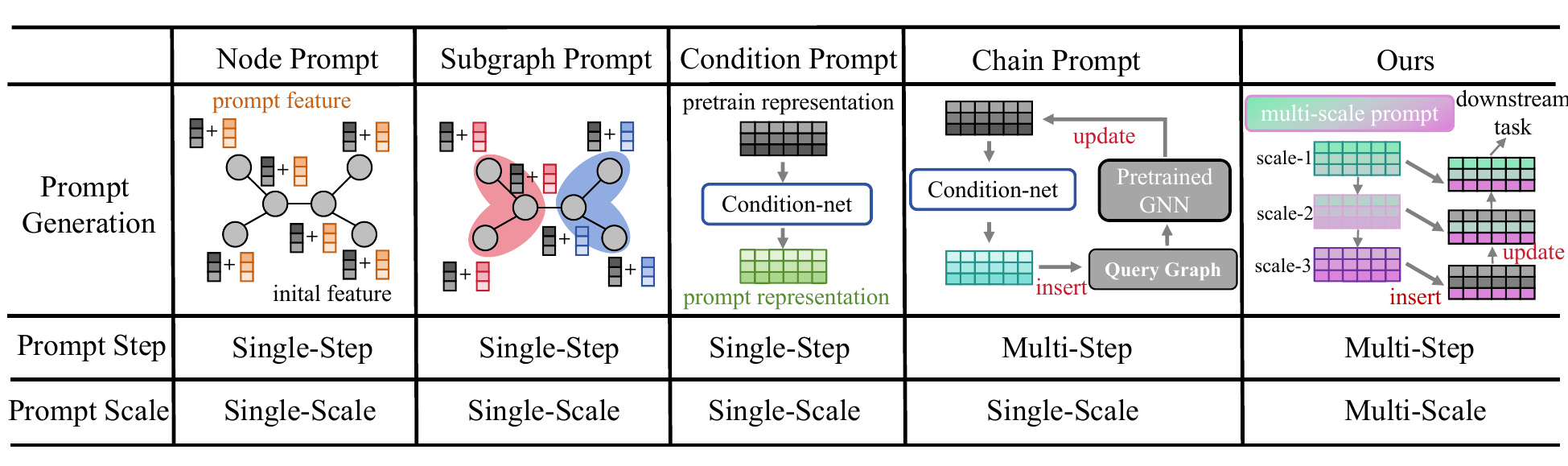} 
\caption{The comparison of graph prompt tuning methods in the existing studies.}
\label{prompt-compare}
\end{figure*}

However, these feature-based prompt tuning methods focus exclusively on single granularity (node-level, edge-level, or subgraph-level) during prompt generation, overlooking the intrinsically coexisting multi-scale information from nodes to hierarchical subgraphs in real-world graphs. As illustrated in Figure~\ref{prompt-compare}, prior works synthesize node/subgraph-specific prompt features using learnable basis vectors or a condition network. GCOT is the current state-of-the-art method designed with text-free multi-step refinement prompts for graph chain of thought architectures, but it still remains confined to node-level granularity at each step. This means that only a single perspective is considered in the prompt generation, fundamentally limiting prompt diversity. 

To overcome the singularity of prompt granularity in existing methods, in this work, we propose a Multi-Scale Graph Chain-of-Thought (MSGCOT) prompt tuning framework. Achieving this design requires solving two core challenges.

First, how to construct multi-granularity information for prompt generation? In hierarchical graph representation learning~\cite{hierarchical}, the original graph is progressively partitioned into smaller subgraphs~\cite{metis,diffpool} by merging nodes or edges to extract multi-scale features. However, these methods are typically designed under supervised settings and ultimately aim to learn a graph representation for solving graph-level tasks. In graph prompt learning, our objective is to obtain a general multi-scale representation that can be adapted to various downstream tasks. Therefore, instead of directly applying the multi-scale representation to downstream tasks, we treat them as an intermediate basis vector pool to enrich node-level prompts with multi-scale features. Specifically, we design a lightweight coarsening network with a low-rank matrix architecture to learn hierarchical coarsened representations. These representations serve as a multi-scale basis vector pool for generating node feature prompts infused with diverse structural granularities.

Second, how to integrate multi-scale information into multi-step reasoning? While the multi-scale basis vector pool provides rich hierarchical features, direct aggregation of all granularities in a single step could cause feature interference and suboptimal prompt generation. Inspired by Chain-of-Thought (CoT) approaches~\cite{cot,cot2}, we find that multi-step progressive prompt optimization is a superior solution. In the NLP domain, CoT requires manually designed textual templates to guide the model's reasoning. For non-textual graph data, we innovatively treat hierarchical coarsened representations as structured thought—from global topology to local details. These multi-granularity representations serve as progressively detailed ``textual examples" in CoT, thereby replacing the guiding role of text. We further propose a backtracking-based progressive prompt optimization strategy: at each reasoning step, the pre-trained embedding is iteratively refined by integrating features of specific granularities, achieving a coarse-to-fine reasoning chain. This process mirrors human cognitive refinement, similar to NLP's iterative text refinement for enhanced answer accuracy. The final generated prompts not only retain the structural outline of coarse granularity but also incorporate discriminative features from fine granularity, enabling more precise capture of hierarchical semantics compared to traditional single-granularity methods.

The main contributions of this work are summarized as follows:
\begin{itemize}
\item We propose the first graph chain of thought framework that integrates multi-granularity information, overcoming the single-granularity limitation of existing methods.
\item We simulate human cognition from coarse-to-fine granularity by designing a low-rank coarsening network for multi-scale feature extraction and a backtracking prompt mechanism for progressive prompt generation.
\item Extensive experiments on eight benchmark datasets for node and graph classification demonstrate that our multi-step, multi-granularity prompting framework outperforms state-of-the-art single-granularity methods.
\end{itemize}

\section{Related Work}
\subsection{Graph Pre-training}
In recent years, graph pre-training have attracted extensive research due to their ability to operate without labeled data~\cite{pretrain,linkpred}. These methods are primarily categorized into contrastive learning~\cite{dgi,graphcl} and generative learning~\cite{graphmae,discrepancyMAE}. Contrastive learning methods construct multiple views by sampling positive and negative samples, then maximizing the consistency between positive samples. Generative learning methods pre-train encoders by reconstructing node features or graph structures. These approaches employ different pre-training objectives to transfer pre-trained knowledge to various downstream tasks. However, due to the significant gap between downstream tasks and pre-training objectives~\cite{allinone,GPL_study}, the performance on downstream tasks may be compromised.

\subsection{Graph Prompt-Tuning}
Recent advances in graph prompt learning have sought to bridge the gap between pre-training and downstream tasks by introducing task-specific prompts. Early works, such as GPPT~\cite{gppt} and GraphPrompt~\cite{graphprompt}, reformulate downstream tasks as link prediction or subgraph similarity tasks. All-in-One~\cite{allinone} unified various downstream tasks as graph-level tasks. Feature-based prompt method GPF+~\cite{GPF} and SUPT~\cite{SUPT} introduced feature-space prompting and subgraph prompt, where learnable vectors modify input. EdgePrompt~\cite{edgeprompt} inserts edge-level prompts in the aggregation at each level from an edge perspective. ProNOG~\cite{pronog} and DAGPrompT~\cite{dagprompt} accounted for node heterophily in the prompt by preserving neighbourhood similarity. The aforementioned methods all belong to single-step prompt tuning approaches. GCOT~\cite{gcot} introduces the chain of thought to generate multi-step prompts. However, these methods only focus on single-granularity information during prompt generation, ignoring the rich multi-granularity features that co-exist in real-world graphs.

\subsection{Graph Coarsening}
Graph coarsening generates simplified graphs by merging nodes or edges. Hierarchical graph coarsening techniques sample a hierarchical learning structure to obtain multiple subgraphs at different scales. Traditional methods compress the graph through spectral clustering~\cite{spectral}, non-negative matrix factorisation~\cite{non-negative_matrix}. In recent years, learnable graph coarsening techniques~\cite{diffpool,hierarchical} progressively aggregate the nodes into coarser graphs through learnable assignment matrices, while preserving important structures, and ultimately generating compact graph-level representations. Graph coarsening naturally provides a hierarchical abstraction mechanism for us to capture multi-scale information in graph prompt learning and learn multi-scale prompts.

\section{Preliminaries}

\para{Notions.}We denote a graph as $G=(\mathcal{V},\mathcal{E})$, where $\mathcal{V}=\{v_1,v_2,\cdots,v_n\}$ represents the set of the nodes and $\mathcal{E} \subseteq \mathcal{V} \times \mathcal{V}$ represents the edge set. Let $N$ represent the total number of nodes. Node features are usually represented by the matrix $\mathbf{X} \in  \mathbb{R}^{N \times F}$, where each row $\mathbf{x}_{i}$ corresponds to the node $i$'s $F$-dimensional feature vector. Let $A \in \mathbb{R}^{N \times N}$ denote the adjacency matrix. Its element $A_{ij}=1$ if there exists an edge between node $i$ and $j$, and otherwise, $A_{ij}=0$.  

\para{Graph Neural Networks.}
Most Graph Neural Networks operate under a message-passing paradigm~\cite{gcn}. Within this scheme, the representation $\mathbf{h}_i$ of each node $i$ is updated iteratively in each layer. This update involves collecting messages from the node's neighborhood, denoted $\mathcal{N}(i)$. Specifically, the representation $\mathbf{h}^{l}_i$ at layer $l$ is computed by first applying an aggregation function $\operatorname{AGGR}(\cdot)$ to the representations from layer $(l-1)$ within $\mathcal{N}(i)$, followed by an $\operatorname{UPDATE}(\cdot)$ function:
	\begin{align}
		&\mathbf{\tilde{h}}^{l}_{i} = \operatorname{AGGR}^{(l)}(\{\mathbf{h}^{l-1}_{j} : j \in \mathcal{N}(i)\}), \\
		&\mathbf{h}^{l}_{i} = \operatorname{UPDATE}^{(l)}(\mathbf{\tilde{h}}^{l}_{i}, \mathbf{h}^{l-1}_{i}).
	\end{align}
In the subsequent sections, to simplify the expression, we denote the encoding process of GNN as follows: 
\begin{equation}
    \mathbf{H}=\operatorname{GNN}(\mathbf{X},\mathbf{A}),
\end{equation}
where $\mathbf{H}$ denotes the final embedding after encoding, and $\operatorname{GNN}(\cdot)$ can conform to any encoder of the message passing paradigm, including but not limited to GCN, GAT, GraphSAGE, and other variants that share this fundamental computational framework.
\section{Proposed Method}

\begin{figure*}[htbp]
\centering   
\includegraphics[width=0.98\textwidth]{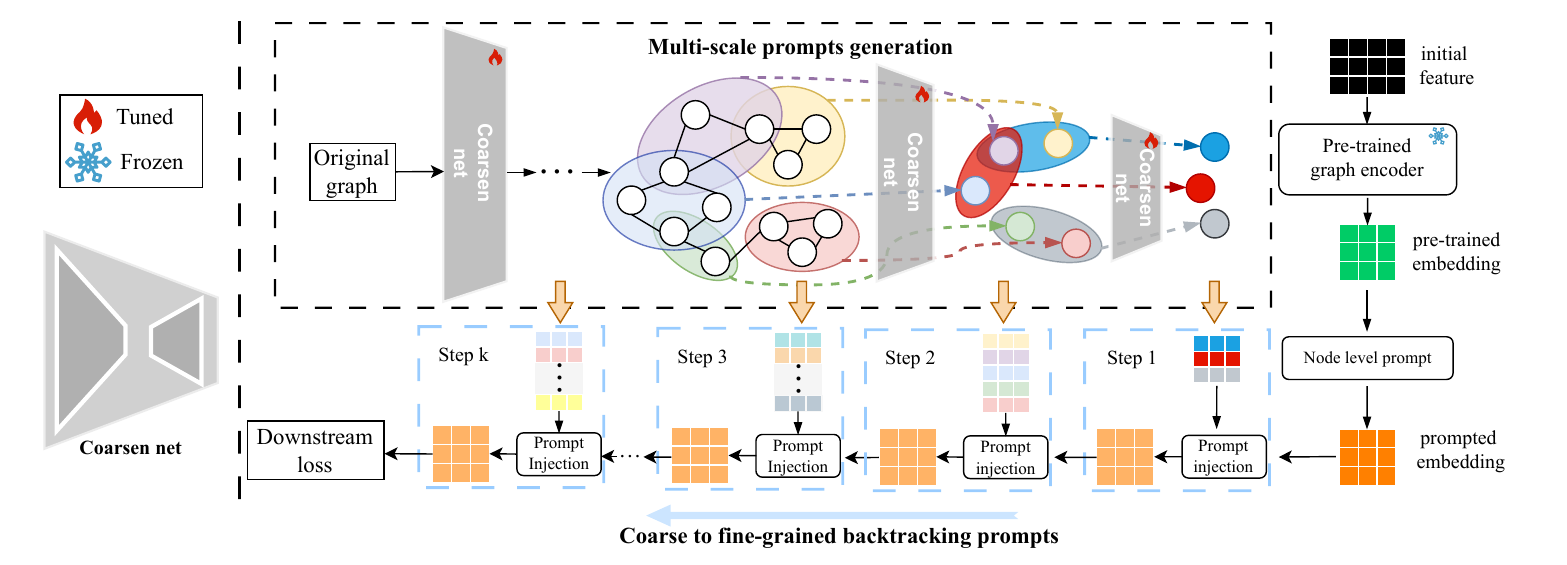} 
\caption{The framework of the proposed MSGCOT.}
\label{Fig.model}
\end{figure*}


\subsection{Overall Framework}
Our objective is to develop a multi-step graph prompt framework that incorporates multi-scale information, as illustrated in Figure \ref{Fig.model}. Specifically, in each reasoning step, the graph coarsening network constructs multi-scale coarsened representations that encompass both coarse-grained and fine-grained information. These coarsened representations as thoughts form a hierarchical basis vector pool for prompt generation. Subsequently, a backtracking-based prompt mechanism is employed to incrementally integrate prompts from coarse to fine granularity, generating hierarchical prompts. Finally, the refined representation is optimized via downstream task loss.

\subsection{Multi-scale Prompt}

\para{Node Level Prompt.}
During the prompt-tuning phase, the parameters of the pre-trained model remain frozen. Using the frozen embeddings directly as input to the coarsening network would limit the model's ability to adapt to downstream tasks. To ensure that the input can be adjusted during training while preserving hierarchical information when extracting multi-scale features, we first generate node-specific prompts for each node. Similar to GCOT~\cite{gcot}, we employ a conditional network along with the initial pre-trained embeddings to produce node-level prompt vectors $P_x$, which serve as task-specific multi-scale features.
\begin{equation}
\mathbf{P_x} = \operatorname{CONDNET}(\mathbf{H}),\\
\end{equation}
where CONDNET is a conditional network by the parameter $\phi$. The conditional network can be viewed as a hypernetwork~\cite{conditional}, i.e., a lightweight auxiliary network. To enhance task-relevant features while suppressing irrelevant noise in the pre-trained representations. Then we use the prompt to modify the original features and obtain prompted embeddings via a pre-trained encoder:
\begin{equation}
    \hat{\mathbf{H}}=\operatorname{GNN}(\mathbf{X} \odot \mathbf{P_x}, \mathbf{A}),
\end{equation}
The node representation after prompting indicates that it contains task adaptation information. The prompted embedding serves a dual role: as an augmented input to the coarsening network and as the initial state of the inference chain.

\para{Multi-scale Thought Construction.}
To inject multiple levels of information into the prompting process, rather than relying only on node iteration updates. We are incrementally updating prompts by constructing multi-scale thoughts so that the prompts contain information at multiple granularities. In traditional CoT~\cite{cot}, thought is defined as the instruction under each reasoning step. Here, we use coarsened representations at different scales to analogise thought under different stages. 

For thought construction, we need to generate a node assignment matrix for each scale. Regarding the implementation of the assignment matrix, there are parametric and non-parametric approaches in earlier work~\cite{metis,diffpool}. In this work, we design a lightweight coarsening network to implement a soft coarsening strategy to generate the assignment matrix $\mathbf{S}^{l}$:
\begin{equation}
    \mathbf{S}^{l} =\operatorname{Softmax}( W^{l}_{up}(\sigma(W^l_{down}\mathbf{T}^{l-1}))),\mathbf{T}^{0}=\mathbf{\hat{H}},
\end{equation}
where the coarsened network consists of a $W_{down} \in \mathbb{R}^{d \times r}$ and $W_{up} \in \mathbb{R}^{r\times C^l}$ $(r \ll d)$. $\sigma$ is the ELU activation function. Here we implement it using the idea of low-rank decomposition~\cite{low_rank}. The number of parameters of the improved coarsening network is much smaller than using an MLP directly. It's consistent with the low resource requirement in prompt learning. After obtaining the assignment matrix, we compute the node representations under coarser granularity:
\begin{equation}
    \mathbf{T}^{l} = {\mathbf{S}^{l}}^{T}\mathbf{T}^{l-1},\\
\end{equation}
where $\mathbf{T}^l \in \mathbb{R}^{n^l \times d}$ denotes the coarsened node representation. Repeating the above operation several times, we obtain a hierarchical thought pool with different levels. These representations serve as base vectors representing different granularities to guide the generation of multi-scale prompts. Instead of using randomly initialised basis vectors directly in earlier methods~\cite{GPF,SUPT,edgeprompt}, here we construct basis vectors incorporating specific scale information.
\begin{equation}
     \mathbf{T} = [\mathbf{T}^l,\mathbf{T}^{l-1},\cdots,\mathbf{T}^1],
\end{equation}
The hierarchical thought pool contains diverse levels and granularities of information, and pre-learned multi-scale thoughts can be used as additional instructions for each reasoning step to generate prompts related to different granularities of information.

\para{Coarse to Fine-grained Backtracking Prompts.}
COT breaks down a complex problem into multiple steps and obtains a more accurate answer by progressively refining the manually designed text. Although GCOT also adopts multi-step prompt generation, the generation of prompts at each step relies on the optimization of the node's features, which is limited to a single granularity. This resembles repeatedly adjusting the same feature dimensions without introducing new perspectives, ultimately constraining prompt semantic diversity. To overcome this, we propose a granularity-aware backtracking mechanism that generates prompts step by step from coarse-grained to fine-grained using multi-granularity basis vectors. This simulates the progressive refinement of textual inputs in large-language-model Q\&A. For each step, the prompts are generated as follows:
\begin{equation}
        \mathbf{p}_i^{l+1} = \sum_j^{C^l} \alpha_{ij}^{l+1} \mathbf{t}_j^{l},a_{ij}^{l+1}=\frac{exp({\mathbf{t}^{l}_j}\hat{\mathbf{h}}_i^l)}{\sum^l_kexp(\mathbf{t}^{l}_k\hat{\mathbf{h}}_i^l)},
\end{equation}
where $\mathbf{p}^{l+1}_i$ denotes the node prompts computed from the thoughts in the hierarchical thought pool, $a_{i,j}^{l+1}$ computes the attention coefficient of the node-level prompts concerning the coarse-grained thoughts, and $\hat{\mathbf{h}}_i^l$ denotes the vector of prompts for the previous step of node $i$, where $\hat{\mathbf{h}}_i^0=\hat{\mathbf{h}}_i$. Next, we inject the prompts with multi-scale information to update the node representation:
\begin{equation}
    \hat{\mathbf{h}}_i^{l+1} = \hat{\mathbf{h}}^l_i+\mathbf{p}^{l+1}_i,
\end{equation}
Based on the constructed multi-layer thought pool, we can use its multi-granularity information to update the representation incrementally. However, the addition of multi-level coarse-grained information may cause the node's unique information to be lost; therefore, we introduce an external cosine reconstruction loss~\cite{graphmae} to constrain the consistency of the node representations with the pre-trained embeddings after prompting. 
\begin{equation}
    \mathcal{L}_{r} = \sum_i^N\frac{1}{N}(1-\frac{\hat{\mathbf{h}_i}\cdot\mathbf{h}_i}{||\hat{\mathbf{h}_i}||\cdot||\mathbf{h}_i||})^ \gamma, \gamma\geq1,
\end{equation}
where $\gamma$ denotes the scaling factor employed to adjust feature reconstruction weights, with parameter settings adhering to prior work. This loss constrains node representations, preventing node-level information loss arising from the introduction of coarse-grained prompts following multi-level prompting.

\subsection{Prompt Tuning}
In the prompt tuning phase, we follow the formulation of ~\cite{graphprompt}, which defines the task within a subgraph similarity framework. Accordingly, our loss function remains consistent with this design and is specified as follows:
\begin{equation}
    \mathcal{L}_{ds}=-\sum_{l=0}^L\sum_{(x_i, y_i) \in \mathcal{D^{\text{train}}}} \ln \frac{\exp ( \text{sim}(\hat{\mathbf{h}}_{x_i}, \hat{\mathbf{h}}_{y_i})/\tau)} {\sum_{c \in \mathcal{Y}} \exp ( \text{sim}(\hat{\mathbf{h}}_{x_i}, \hat{\mathbf{h}}_c)/ \tau)},
\end{equation}
where $\tau$ denotes the temperature constant and sim($\cdot$) denotes the cosine similarity. $\hat{\mathbf{h}}_{x_i}$ denotes the final node embedding or graph embedding, and $\hat{\mathbf{h}}_c$ the prototype embedding for class $c$, obtained from the class mean of the labelled nodes or graphs of that class. Thus, the final loss is:
\begin{equation}
    \mathcal{L}_{final} = \mathcal{L}_{ds} + \alpha\mathcal{L}_r,
\end{equation}
where $\alpha$ denotes the weight of the reconstruction loss. For the node classification task, it is important to maintain the original node-level information while injecting multi-granularity information, while for graph classification, the focus is more on coarse-grained graph-level information. During prompt tuning, only the weights of the lightweight coarsened network and the weights of the node-level conditional network are updated.

\subsection{Complexity Analysis}
The computational complexity of MSGCOT primarily stems from two key components: multi-scale information extraction and a chain of graph prompt generation. Let $N$ denote the total number of nodes, $C^l$ represent the coarsened nodes at layer $l$, $L$ be the number of coarsening layers, $d$ be the hidden dimension, and $r$ be the low-rank dimension. For the coarsening network, each layer's time complexity is $O(C^{l-1}(dr + rC^l))$ where $C^0 = N$. With a coarsening ratio $c = C^l/C^{l-1} < 1$ ( $C^l = c^l N$), the total complexity across $L$ layers becomes $\sum_{l=1}^{L} O(c^{l-1} N (dr + r c^l N))$, which by geometric series convergence simplifies to $O\left(\frac{dr(1-c^L)N}{1-c} + \frac{rc(1-c^{2L})N^2}{1-c^2}\right)$. The prompt generation phase contributes $\sum_{l=1}^{L}O(N^2c^ld) = O(\frac{N^2c(1-c^L)d}{1-c})$. Experimentally, setting $c \in [0,0.2]$ makes $c^{L}$ and $c^{2L}$ negligible, reducing the total complexity to $O\left(\frac{drN}{1-c} + \frac{rcN^2}{1-c^2}+\frac{dcN^2}{1-c}\right)$. For small $c (\leq 0.2)$, the denominators $(1-c)$ and $(1-c^2)$ approach 1, further simplifying to $O(drN + (r+d)cN^2)$. When $cN \ll N$, the overall complexity becomes nearly linear, ensuring computational feasibility and efficiency. Please see the Sections \ref{time and memory} for more details about time-efficiency and parametric experiments.

\section{Experiments}
\begin{table*}[htbp]
\caption{Performance comparison on node and graph classification tasks. The best and second-best results are highlighted in bold and underlined, respectively.}
\centering
\renewcommand{\arraystretch}{1.3} 
\begin{tabular}{l|cccc|cccc}
    \midrule
 & \multicolumn{4}{c|}{\textbf{Node Classification}} & \multicolumn{4}{c}{\textbf{Graph Classification}} \\
\cline{2-9}

 & \textbf{Cora} & \textbf{Citeseer} & \textbf{Pubmed} & \textbf{Photo} & \textbf{MUTAG} & \textbf{COX2} & \textbf{BZR} & \textbf{PROTEINS} \\
    \midrule
GCN & 33.33$\pm$13.68 & 26.28$\pm$8.79 & 52.26$\pm$8.95 & 60.79$\pm$11.96 & 50.94$\pm$7.45 & 50.05$\pm$4.79 & 49.98$\pm$5.85 & 52.38$\pm$6.88 \\
GAT & 31.24$\pm$15.05 & 29.20$\pm$8.85 & 47.62$\pm$9.40 & 50.74$\pm$13.78 & 48.87$\pm$8.54 & 49.23$\pm$7.40 & 50.37$\pm$8.40 & 53.41$\pm$8.56 \\
    \midrule

LP & 56.51$\pm$13.48 & 43.52$\pm$9.34 & 53.98$\pm$7.86 & 62.43$\pm$9.32 & 59.13$\pm$14.54 & 51.28$\pm$4.56 & 51.37$\pm$9.06 & 52.38$\pm$5.91 \\
DGI/Infograph & 55.69$\pm$12.32 & 45.64$\pm$9.32 & 54.38$\pm$10.26 & 63.98$\pm$9.60 & 58.92$\pm$16.34 & 51.58$\pm$14.68 & 51.28$\pm$9.84 & 53.64$\pm$8.02 \\
GraphCL & 55.35$\pm$10.72 & 45.64$\pm$10.03 & 53.54$\pm$8.47 & 64.56$\pm$9.85 & 59.02$\pm$14.13 & 51.64$\pm$14.37 & 51.53$\pm$13.58 & 54.23$\pm$7.65 \\
    \midrule

All-in-One & 51.74$\pm$12.54 & 42.39$\pm$8.27 & 62.93$\pm$10.75 & 66.06$\pm$10.63 & 58.58$\pm$14.25 & 53.33$\pm$14.12 & 48.95$\pm$13.68 & 53.52$\pm$8.12 \\
GPF & 58.45$\pm$13.35 & 46.22$\pm$8.51 & \underline{63.40$\pm$9.80} & \underline{67.37$\pm$10.42} & \underline{62.11$\pm$14.95} & 53.96$\pm$14.40 & 48.27$\pm$12.76 & 55.31$\pm$8.96 \\
GPF+ & 57.07$\pm$14.14 & 44.03$\pm$8.74 & 56.87$\pm$10.88 & 65.77$\pm$10.54 & 58.63$\pm$14.92 & 54.60$\pm$13.57 & 50.73$\pm$13.66 & 54.44$\pm$9.08 \\
SUPT & 55.70$\pm$13.55 & 43.85$\pm$8.65 & 57.86$\pm$11.31 & 63.45$\pm$10.30 & 60.31$\pm$16.44 & 54.88$\pm$13.53 & 48.38$\pm$10.41 & 54.36$\pm$8.96 \\
GraphPrompt & 58.42$\pm$13.31 & 46.09$\pm$8.42 & 63.31$\pm$9.84 & 67.35$\pm$10.38 & 59.84$\pm$14.66 & 51.44$\pm$4.30 & 51.27$\pm$8.92 & 53.44$\pm$5.99 \\
DAGPrompt & 57.77$\pm$12.56 & 46.23$\pm$4.46 & 60.51$\pm$10.80 & 54.73$\pm$10.83 & 58.99$\pm$14.69 & 55.00$\pm$13.00 & \underline{55.49$\pm$13.74} & \underline{56.22$\pm$10.53} \\
EdgePrompt & 57.14$\pm$14.21 & 44.01$\pm$8.62 & 53.66$\pm$11.86 & 64.90$\pm$10.83 & 62.06$\pm$14.40 & 53.83$\pm$14.66 & 46.94$\pm$12.47 & 55.23$\pm$9.90 \\
EdgePrompt+ & 55.93$\pm$14.24 & 44.05$\pm$8.93 & 57.59$\pm$11.15 & 64.50$\pm$10.08 & 59.74$\pm$15.45 & \underline{55.37$\pm$13.79} & 48.66$\pm$12.03 & 54.93$\pm$9.36 \\
GCOT & \underline{59.54$\pm$13.60} & \underline{48.13$\pm$8.89} & 63.38$\pm$9.98 & 66.98$\pm$10.65 & 60.34$\pm$14.70 & 52.09$\pm$12.34 & 54.45$\pm$16.19 & 55.07$\pm$10.51 \\
    \midrule

\textbf{MSGCOT} & \textbf{62.13$\pm$17.53} & \textbf{49.05$\pm$11.41} & \textbf{64.67$\pm$10.41} & \textbf{68.01$\pm$10.39} & \textbf{63.54$\pm$14.94} & \textbf{73.62$\pm$6.43} & \textbf{69.85$\pm$11.65} & \textbf{57.83$\pm$2.71} \\
    \bottomrule
        \end{tabular}
\label{class}
\end{table*}

\subsection{Experinmental Setting}
\para{Datasets.}
We evaluate the performance of MSGCOT using several datasets from different domains, including citation networks, e-commerce, protein structures, and molecular graphs. For the node classification task, we used three citation network datasets, Cora~\cite{cora}, Citeseer, Pubmed~\cite{citeseerpub}, and the e-commerce co-purchase network Photo~\cite{photo}. For graph classification, we used a protein structure dataset PROTEINS~\cite{protein}, a molecular graph dataset MUTAG, BZR, and COX2~\cite{mutagcoxbzr}. 

\para{Baselines.}
We compare MSGCOT with current state-of-the-art methods. These baselines fall into three categories:(1) Supervised learning. We use a supervised approach on GCN~\cite{gcn}, GAT~\cite{gat} to train on downstream labels. (2) Pre-training + fine-tuning. LP~\cite{linkpred}, GraphCL~\cite{graphcl}, and DGI/InfoGraph~\cite{dgi,infograph} employ a pre-training and fine-tuning strategy, where the model is pre-trained on unlabeled data and then fine-tuned on downstream tasks. (3) Pre-training + Prompt: We uniformly use the link prediction task to pre-train the GNNs, and use graph prompt methods for prompting, including single-step prompt methods All-in-One~\cite{allinone}, GPF, GPF+~\cite{GPF}, SUPT~\cite{SUPT}, GraphPrompt~\cite{graphprompt}, EdgePrompt, EdgePrompt+~\cite{edgeprompt}, DAGPrompT~\cite{dagprompt}, and multi-step prompt method GCOT~\cite{gcot}. 

\para{Implementation details.}
We evaluate our approach on a few-shot node classification and graph classification tasks. For each class, we randomly select 
$k$ nodes or graphs as the training set. For node classification, following the GCOT setup, we randomly sample 1,000 nodes from the remaining nodes as the test set. For graph classification, we use the remaining samples as the test set. We report the model accuracy. To ensure a fair comparison, all prompt-based models adopt link prediction as the pre-training strategy, using GCN as the backbone with a hidden layer size of 256. The number of coarsening layers is set to 2, and the coarsening ratio is selected from {0.01,0.1,0.2,0.3}. The hidden dimension of the low-rank decomposition matrix is set to 8. The prompt constraint weight $\alpha$ is set to 1 for node classification and 0 for graph classification. We conduct 100 random sampling trials, generating one-shot tasks for both node classification and graph classification, and report the mean and variance of the results. The pre-training strategies and hyperparameter settings for other baseline methods follow their original papers. All experiments are conducted on an NVIDIA RTX 4090D GPU with 24GB of memory. Experimental results using GraphCL as a pre-training task are provided in Appendix \ref{pre+graphcl}. Detailed parameter settings can be found in Appendix \ref{hyperparaset}.

\subsection{Performance on One-Shot Classification}
We present the results of node classification and graph classification in Table~\ref{class}, revealing the following findings: (1) MSGCOT consistently outperforms all baseline methods across all settings. On the Cora dataset, MSGCOT achieves a 3.68\% improvement over the second-best method GPF, and demonstrates significant gains compared to the multi-step prompting approach GCOT, highlighting the importance of hierarchical prompts for node classification tasks. (2) For graph classification, MSGCOT achieves substantial improvements of 5-20\% over the strongest baselines. Notably, on the COX2 dataset, it surpasses the second-best method DAGPrompt by 18.62\%, as the multi-scale prompts effectively capture latent substructures. Unlike GCOT, which only incorporates node-level information in multi-step prompts, MSGCOT dynamically adjusts prompt granularity to preserve local structures while enhancing global representations, demonstrating superior performance.

\subsection{Performance on Few-Shot Classification}
\begin{figure}[ht]
\centering
\subfigure[Cora]{\includegraphics[width=0.48\columnwidth]{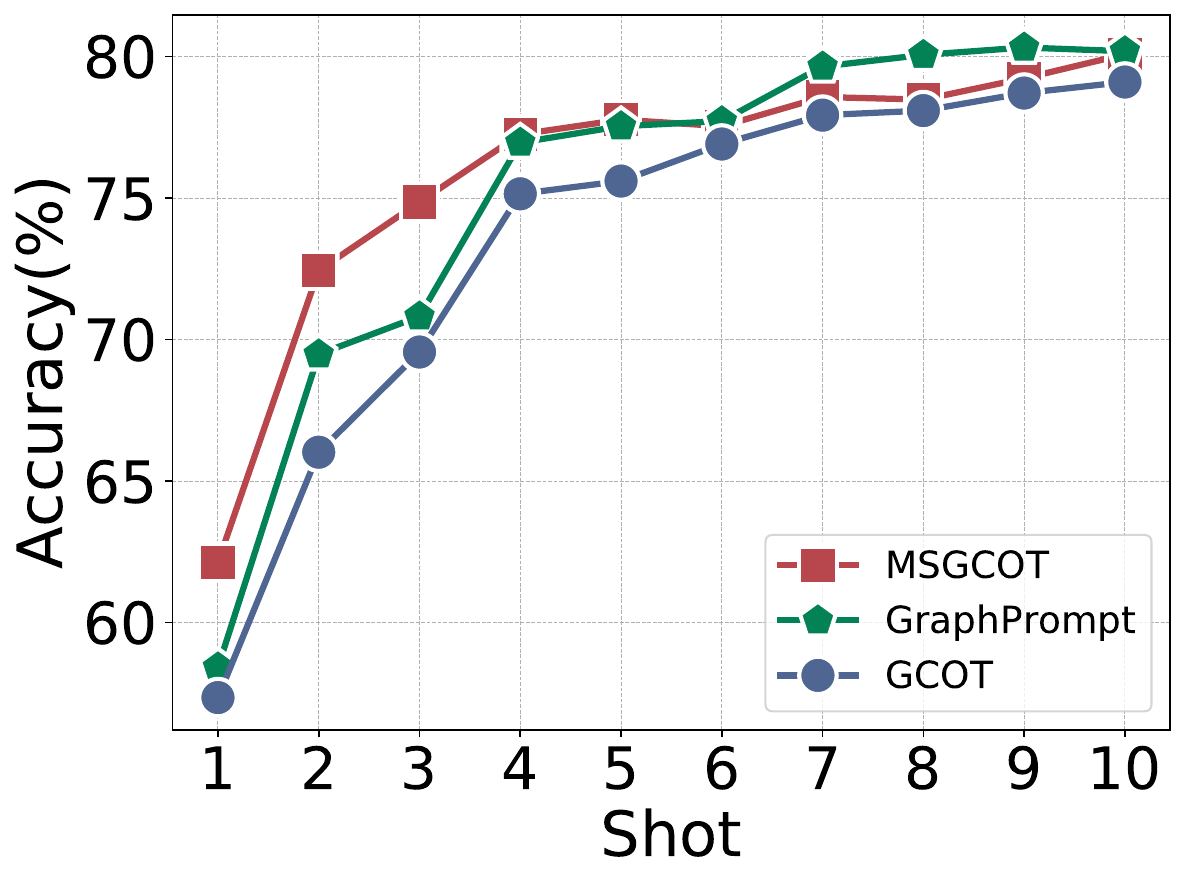}\label{fig:node_shot}}
\subfigure[MUTAG]{\includegraphics[width=0.48\columnwidth]{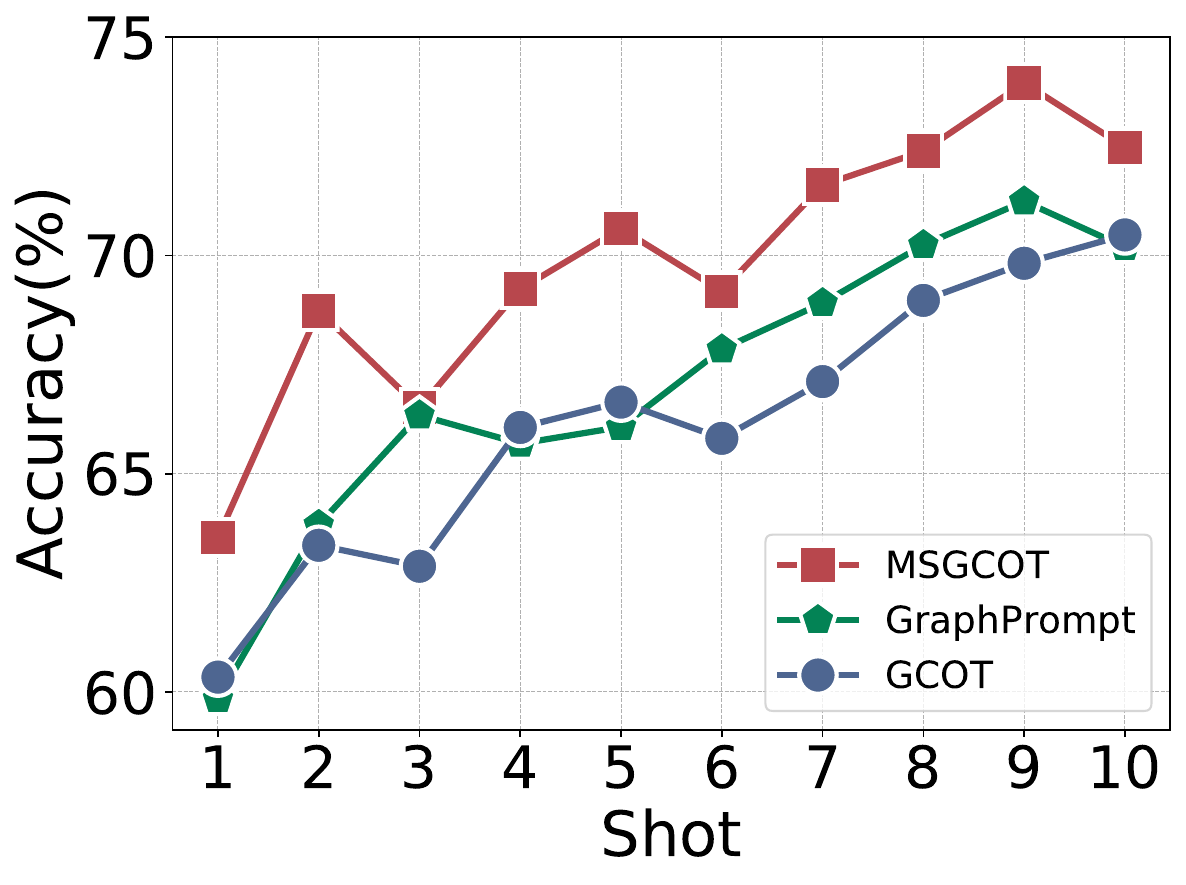}\label{fig:graph_shot}}
\caption{Impact of shots number analysis.}
\label{fig:fewshot}
\end{figure}

To explore the impact of the amount of labelled data on the model performance, we adjust the number of shots. As shown in Figure~\ref {fig:fewshot}, we find that MSGCOT exhibits significant advantages in the few-shot scenario. In the node classification, MSGCOT outperforms the baseline methods by an average of 5-8\% in the 1-3 shot setting, and as the sample size increases to 5-10 shots, the performance of all methods improves, but MSGCOT remains competitive. In the graph classification, MSGCOT demonstrates significant advantages at different shots on MUTAG, confirming the robustness of the method to data scarcity. When the sample size is increased, the performance remains superior compared to the multi-step single-granularity prompt method GCOT. These results validate the effectiveness of the multi-scale prompt in few-shot learning.


\subsection{Ablation study}
To comprehensively evaluate the effectiveness of the multi-scale prompt mechanism in the MSGCOT framework, we conducted systematic ablation studies by comparing MSGCOT with four variants: (1) w/o multi-scale prompt(MSP): removing multi-scale prompts and using only node-level prompts; (2) w/o reconstruction loss(RE): eliminating the reconstruction constraint on multi-level node prompts; (3) w/o trackback (TB): employing a unidirectional process (fine-to-coarse) for prompt addition; and (4) w/o incremental update (IU): retaining only the final prompt.

As shown in Table~\ref{abation}: (1) Multi-scale prompts are crucial for performance. Their removal reduces node classification accuracy by 5.52\% on average and leads to a more pronounced drop in graph classification (17.7\% on average), validating their critical role in capturing hierarchical structural features. (2) The reconstruction loss significantly impacts node classification, while graph classification relies more on global information. (3) The backtracking mechanism is particularly vital for graph classification; performance drops sharply by 12-15\% when using unidirectional prompts, confirming the importance of dynamically adjusting prompt granularity. (4) Progressive prompt updates yield a 2-5\% performance gain, indicating that intermediate-layer prompts carry complementary hierarchical information. These results demonstrate that MSGCOT effectively coordinates features across different granularities, verifying the efficacy of its multi-level, multi-scale prompt design.

\begin{table}[h]
\caption{Performance comparison of different variants}
\centering
    \renewcommand{\arraystretch}{1.3} 
    \setlength\tabcolsep{1.2pt} 
\begin{tabular}{l|cccc}
\midrule
 & Cora & Citeseer & COX2 & MUTAG \\
\midrule
w/o MSP & 56.61$\pm$14.66 & 43.66$\pm$9.03 & 51.49$\pm$4.21  & 60.13$\pm$14.69 \\
w/o RE & 46.98$\pm$15.90 & 39.40$\pm$10.72 & -  & - \\
w/o TB & 59.67$\pm$17.01 & 48.38$\pm$10.51 & 56.33$\pm$12.83  & 62.45$\pm$16.27 \\
w/o IU & 60.61$\pm$16.45 & 46.59$\pm$10.00 & 53.58$\pm$14.59  & 62.10$\pm$14.57 \\
FULL & \textbf{62.13$\pm$17.53} & \textbf{49.05$\pm$11.41} & \textbf{73.62$\pm$6.43} & \textbf{63.54$\pm$14.94} \\
\midrule
\end{tabular}
\label{abation}
\end{table}

\subsection{Hyperparameter Analysis}
\begin{figure}[ht]
\centering
\subfigure[Node classification]{\includegraphics[width=0.48\columnwidth]{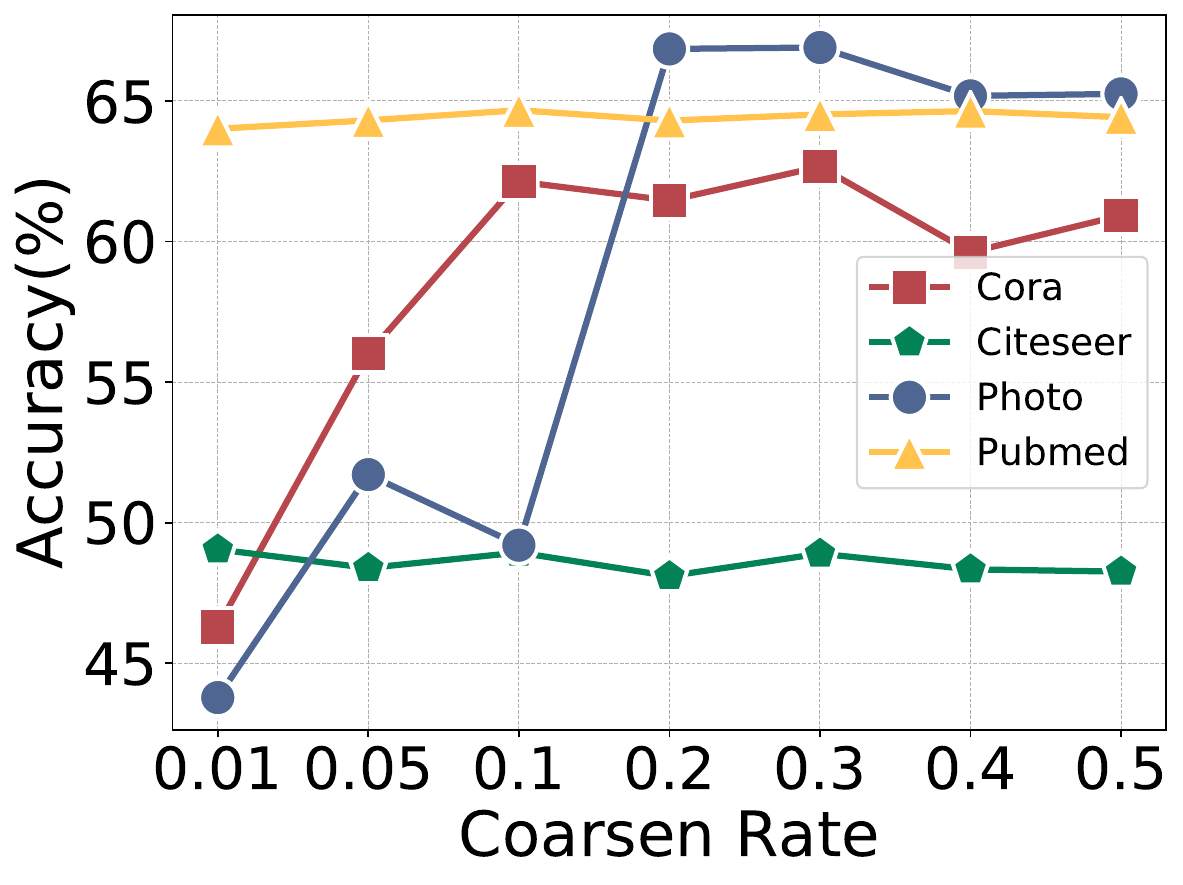}\label{fig:node_rate}}
\subfigure[Graph classification]{\includegraphics[width=0.48\columnwidth]{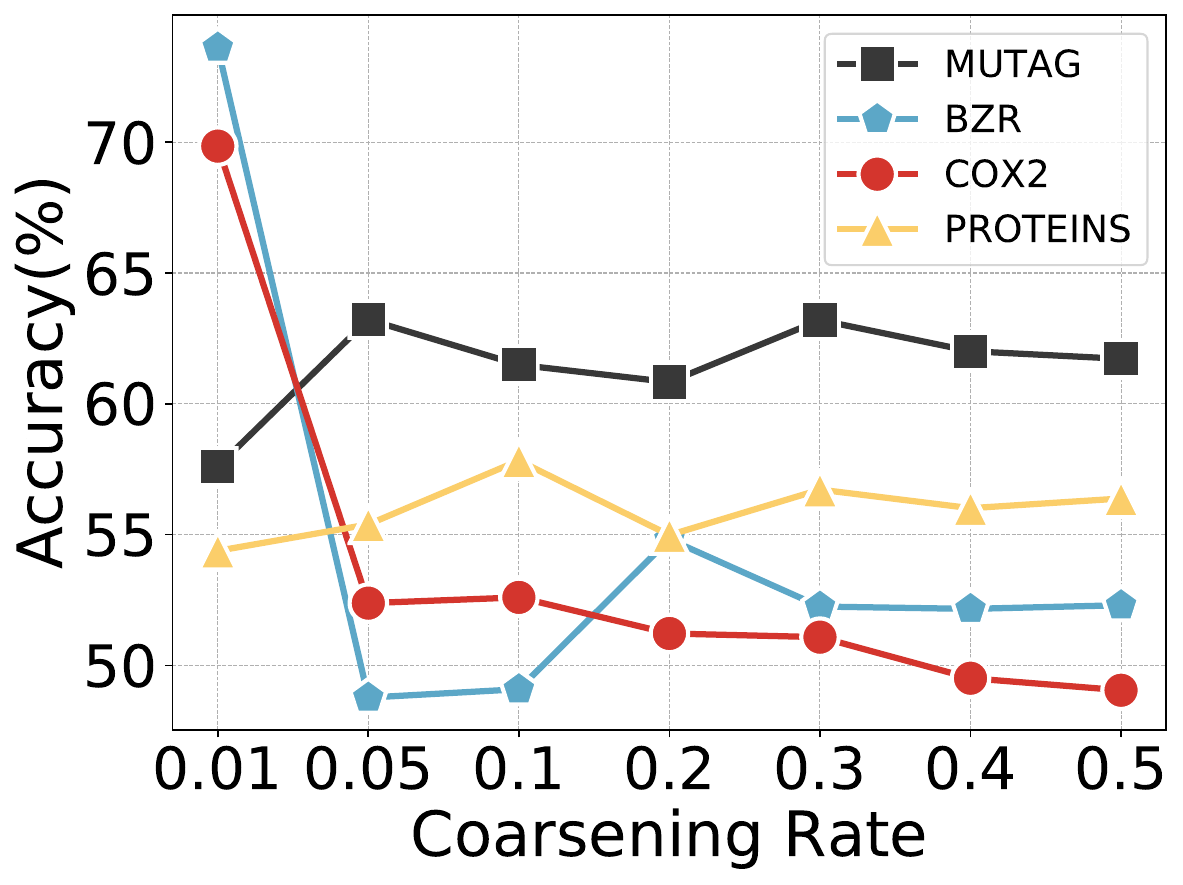}\label{fig:graph_rate}}
\caption{The sensitivity of the coarsening rate.}
\label{fig:rate}
\end{figure}
\para{Coarsening rate.}
We control the number of coarsened nodes via the coarsening rate, where lower values emphasize coarse-grained information. As shown in Figure~\ref{fig:rate}, node-level tasks peak at moderate rates (0.1–0.3), as excessively low rates lose fine-grained details. Graph classification remains robust across a wider range (0.05–0.3), demonstrating stronger robustness to coarsening.

\begin{figure}[ht]
\centering
\subfigure[Node classification]{\includegraphics[width=0.48\columnwidth]{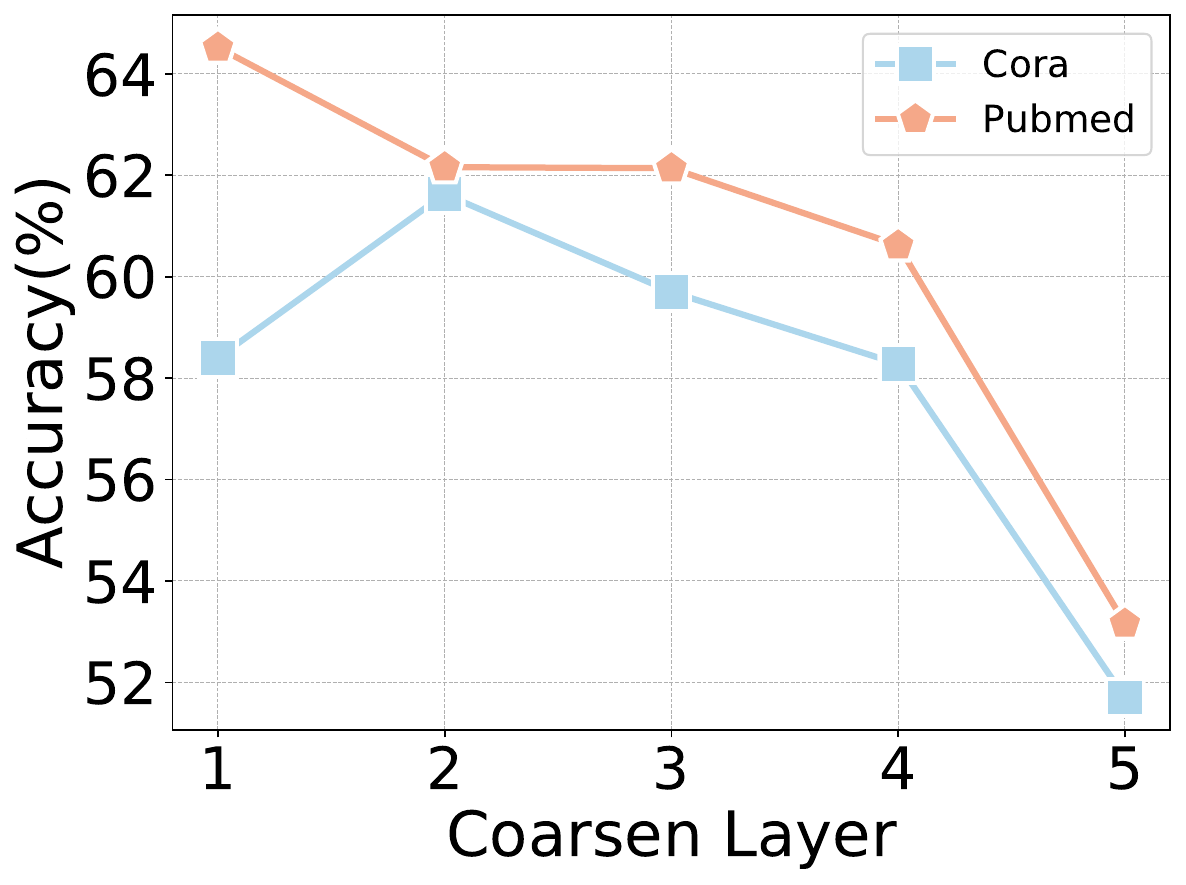}\label{fig:node_l}}
\subfigure[Graph classification]{\includegraphics[width=0.48\columnwidth]{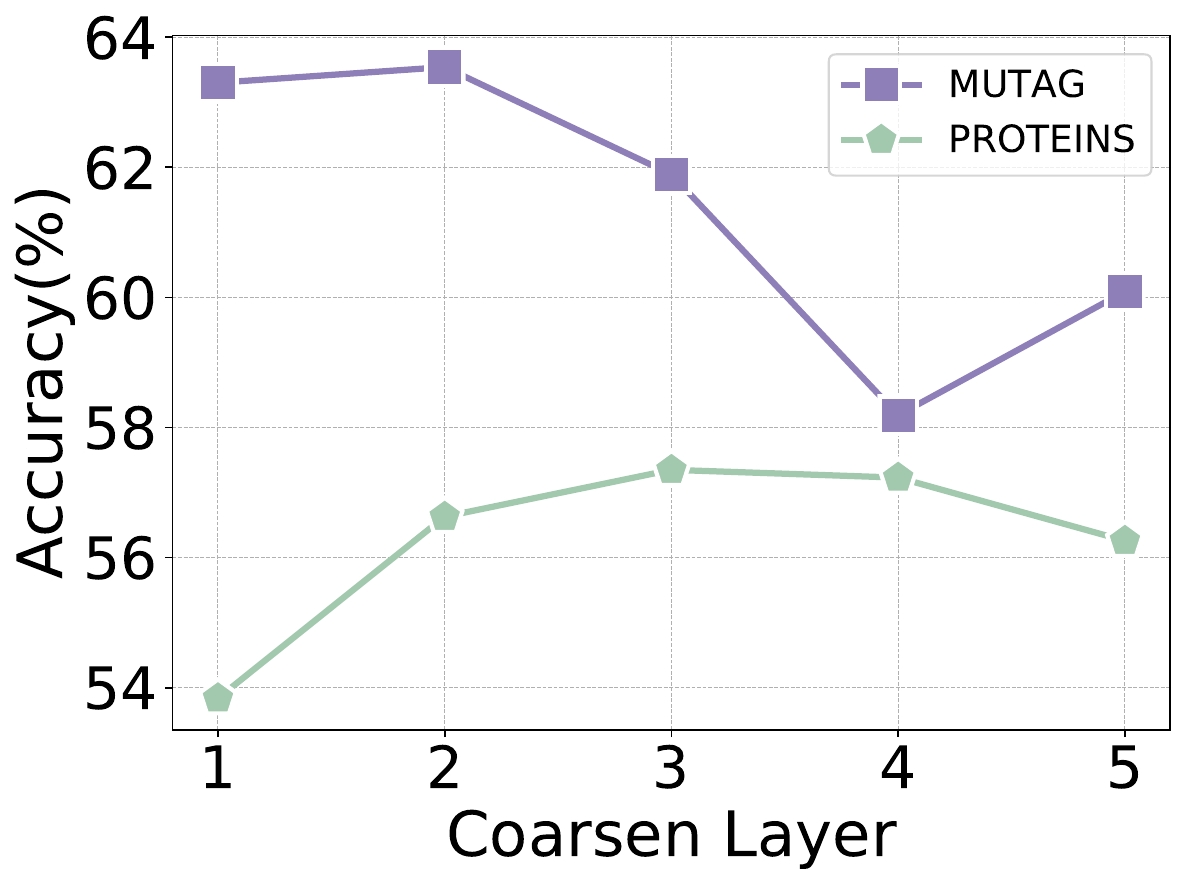}\label{fig:graph_l}}
\caption{The sensitivity of the coarsening layer number.}
\label{fig:layer}
\end{figure}

\para{Coarsening layer.}
The number of coarsening layers defines both the depth of multi-scale prompting and the reasoning steps. At each step, hierarchical prompts of varying granularity are integrated. We analyze the effect of layer count in Figure~\ref{fig:layer} and observe: (1) For node-level tasks, performance initially improves then declines with more layers. On Cora, two layers achieve an optimal balance between node semantics and subgraph structure; excessive layers induce over-smoothing. (2) For graph-level tasks, performance steadily improves with deeper hierarchies. On PROTEINS, performance remains strong at 3–4 layers, indicating enhanced benefits from hierarchical feature extraction. This confirms graph tasks' greater reliance on multi-scale structures and validates our hierarchical prompting mechanism.

\begin{figure}[h]
\centering
\subfigure[Node classification]{\includegraphics[width=0.48\columnwidth]{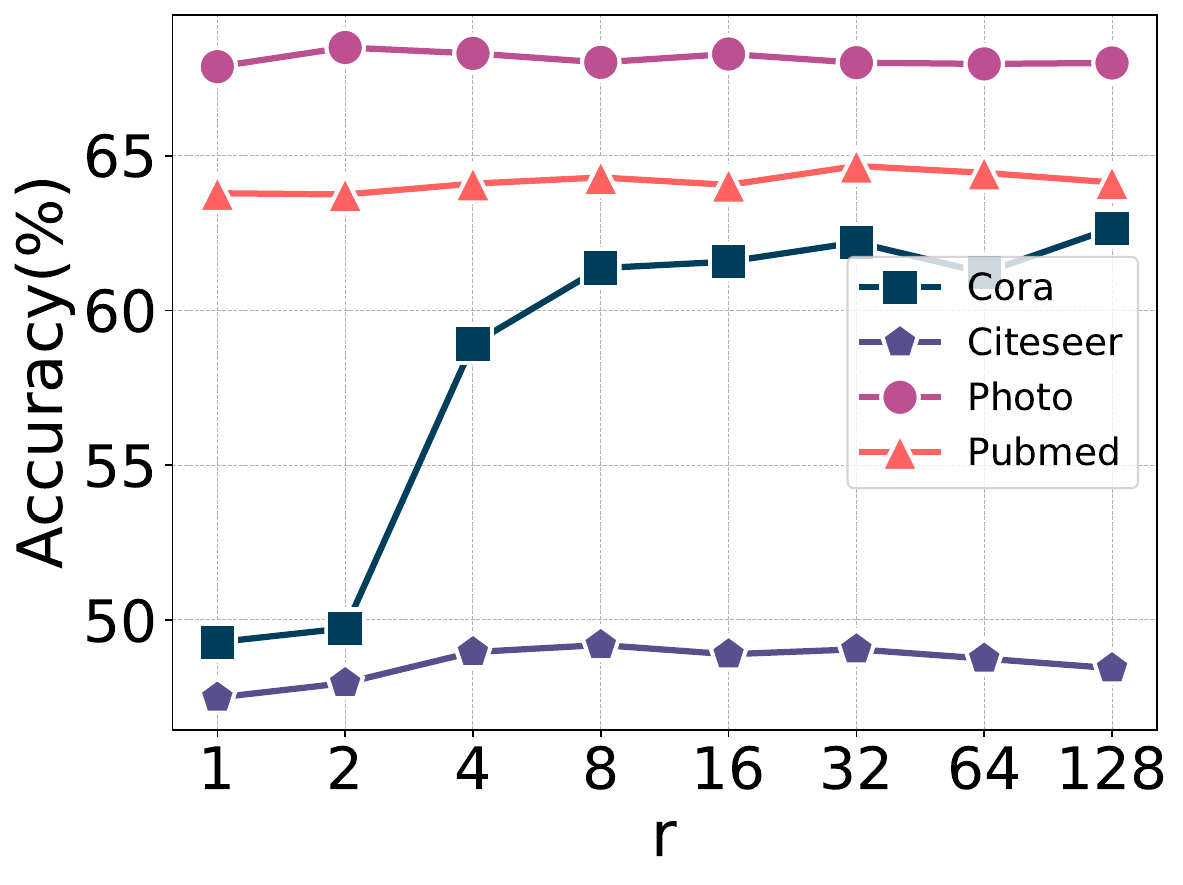}\label{fig:node_rank}}
\subfigure[Graph classification]{\includegraphics[width=0.48\columnwidth]{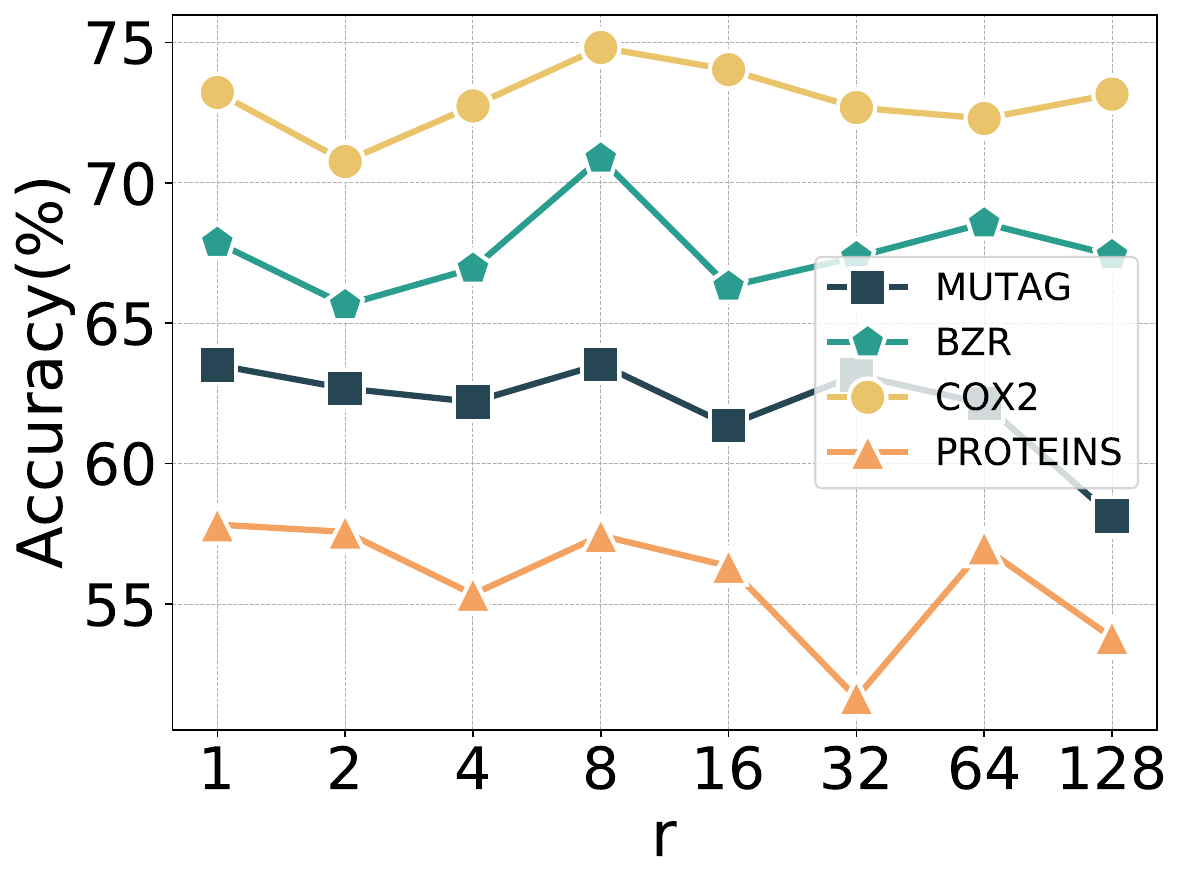}\label{fig:graph_rank}}
\caption{The sensitivity of the hidden dimensions of the coarsening net.}
\label{fig:rank}
\end{figure}

\para{Hidden dimensionality of coarsening net.}
We design an efficient coarsening network to capture multi-scale information and improve generalization by controlling parameter scale. To assess its efficacy, we evaluate performance across different hidden dimensions $r$ (Figure~\ref{fig:rank}). For node classification, accuracy stabilizes when $r>8$, with peak results at $r=8$. Graph classification exhibits greater robustness to dimension changes, achieving strong performance even at $r=1$. A slight overfitting emerges in node tasks when $r>32$. These results confirm that MSGCOT's low-rank design effectively balances efficiency and performance stability, sustaining strong generalization with minimal dimensionality.

\subsection{The Effectiveness of Multi-Scale Prompts}
\begin{figure}[ht]
\centering
\subfigure[Node classification]{\includegraphics[width=0.48\columnwidth]{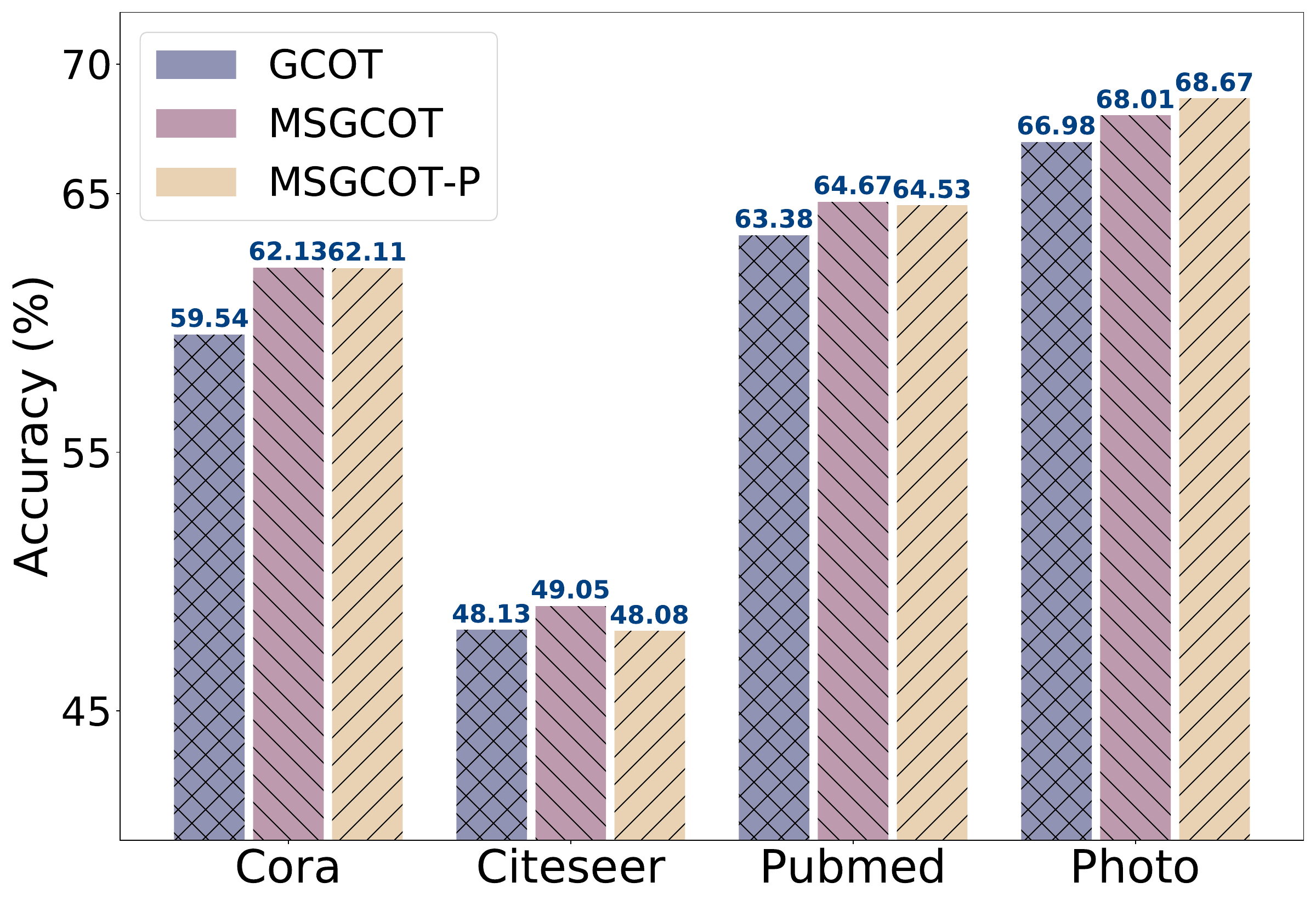}\label{fig:node_sca}}
\subfigure[Graph classification]{\includegraphics[width=0.48\columnwidth]{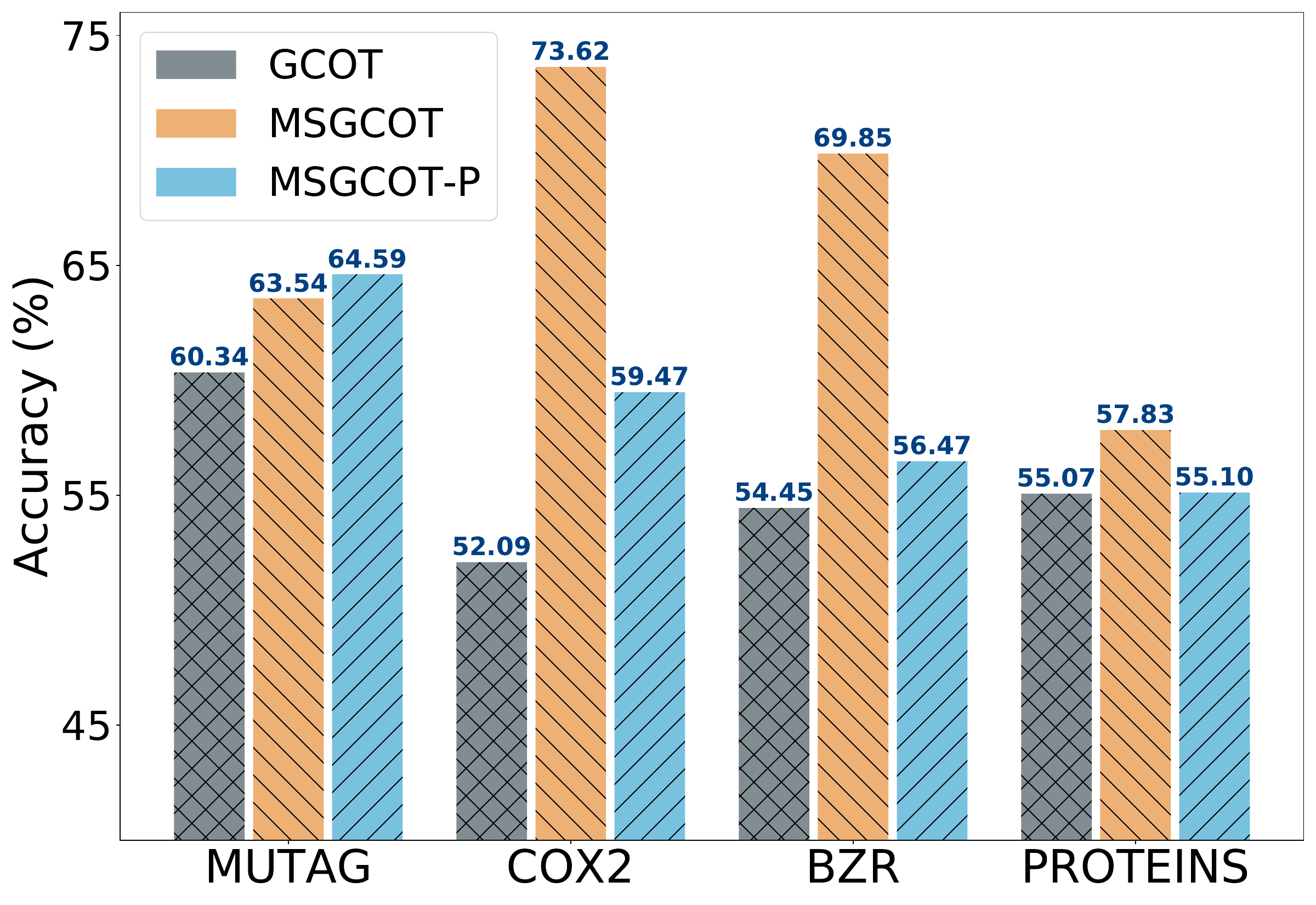}\label{fig:graph_sca}}
\caption{The extensibility analysis of MSGCOT.}
\label{fig:sca}
\end{figure}

To validate the effectiveness of multi-scale prompts, we evaluate a non-parametric variant (MSGCOT-P) that replaces the trainable coarsening network with precomputed multi-scale partitions $[S^1, S^2, \dots, S^l]$ derived from graph coarsening algorithms~\cite{metis}. This eliminates the need for learnable assignment matrices, reducing trainable parameters to only the node-level prompt components. Therefore, this version has smaller training parameters. As shown in Figures \ref{fig:node_sca} and \ref{fig:graph_sca}, MSGCOT-P achieves comparable or superior results to GCOT across all datasets, with notable improvements on Photo (+1.69\%) and MUTAG (+4.25\%). While slightly trailing the parametric MSGCOT in most cases, it outperforms all other baselines, demonstrating that multi-scale reasoning drives performance gains. The strong performance of MSGCOT-P highlights the framework’s adaptability to existing graph partitioning strategies.

\subsection{Parameter and Efficiency Analysis}
\label{time and memory}
\para{The effectiveness of lightweight coarse-grain networks.}
The parameters of MSGCOT primarily come from the coarsening network, which consists of two low-rank decomposed matrices. For each layer, the number of parameters can be calculated as 
$dr + rC^l$. The total parameter count is $Ldr +\sum_{l=1}^LrC^l$. Taking the first layer as an example, we compare the parameter counts with and without the lightweight coarsening network. As shown in Figure 6, node classification nearly reaches its optimal performance at $r=8$, while graph classification achieves strong results even at $r=1$. Here, we set $r=8$ on the node task and r=1 on the graph task. As shown in the Table~\ref{tab:low_rank_comparison}, the low-rank design maintains a stable parameter count of 0.43K-4.71K, whereas the full-rank design requires 43.15K-85.17K. The relative ratio is as low as 0.9\% (COX2) and up to 6.1\% (Cora), validating the efficiency of low-rank decomposition.

\begin{table}[h]
    \caption{Comparison of prompt-tuning training parameters on rank and without low rank.}
    \centering
    \begin{tabular}{l|cccc}
        \toprule
        & \textbf{Cora} & \textbf{Citeseer} & \textbf{MUTAG} & \textbf{COX2} \\
        \midrule
        low rank & 4.21K & 4.71K & 0.43K & 0.45K \\
        w/o low rank & 69.32K & 85.17K & 43.15K & 49.28K \\
        \midrule
        Relative Ratio & 6.1\% & 5.5\% & 1.0\% & 0.9\% \\
        \bottomrule
    \end{tabular}
    \label{tab:low_rank_comparison}
\end{table}

\para{Trainable prompt parameters.}We compare trainable parameters across prompting methods in Table~\ref{tab:parameter_comparison}. MSGCOT maintains competitive performance while substantially reducing parameters: for node classification (10.37K–21.8K), it achieves a 47.1\%–68.3\% reduction compared to GCOT (32.76K). In graph classification (1.17K–5.81K), reductions reach 29.1\%–85.7\% versus GCOT (8.19K), even outperforming GPF+ (5.12K) on MUTAG. This efficiency stems from MSGCOT’s low-rank design (r=8 for node tasks, r=1 for graph tasks) and task-adaptive coarsening. On average, parameter reductions are 53.6\% (node) and 63.2\% (graph), highlighting its ability to balance complexity and performance.
\begin{table}[h]
    \caption{Comparisons of prompt-tuning training parameters (K) on node and graph classification tasks.}
    \centering
    \begin{tabular}{l|cccc}
        \toprule
        \textbf{Dataset} & \textbf{GPF+} & \textbf{EdgePrompt+} & \textbf{GCOT} & \textbf{MSGCOT} \\
        \midrule

        Cora & 5.12 & 10.24 & 32.76 & 10.37 \\
        Citeseer & 5.12 & 10.24 & 32.76 & 11.01 \\
        Pubmed & 5.12 & 10.24 & 32.76 & 21.80 \\
        Photo & 5.12 & 10.24 & 32.76 & 14.81 \\
        \midrule
   
        MUTAG & 5.12 & 10.24 & 8.19 & 1.20 \\
        BZR & 5.12 & 10.24 & 8.19 & 1.17 \\
        COX2 & 5.12 & 10.24 & 8.19 & 1.24 \\
        PROTEINS & 5.12 & 10.24 & 8.19 & 5.81 \\
        \bottomrule
    \end{tabular}
    \label{tab:parameter_comparison}
\end{table}

\para{Running times.}
Table~\ref{tab:time_comparison} compares the per-epoch running times of different prompt tuning methods. MSGCOT shows clear efficiency benefits across task types. For node classification, its runtime is slightly higher than GPF+ and EdgePrompt+ but comparable to GCOT. For graph classification, MSGCOT averages 0.137 seconds per epoch—a 34.8\% reduction over GCOT—while maintaining higher accuracy, attributable to its low coarsening ratio and reduced hidden dimension. Although multi-step prompting adds complexity, MSGCOT's overhead remains manageable.
\begin{table}[h]
    \caption{Comparisons of prompt-tuning times (seconds) on node and graph classification tasks.}
    \centering
    \begin{tabular}{l|cccc}
        \toprule
        \textbf{Dataset} & \textbf{GPF+} & \textbf{EdgePrompt+} & \textbf{GCOT} & \textbf{MSGCOT} \\
        \midrule
        Cora & 0.225 & 0.214 & 0.246 & 0.250 \\
        Citeseer & 0.234 & 0.214 & 0.250 & 0.248 \\
        Pubmed & 0.224 & 0.210 & 0.272 & 0.282 \\
        Photo & 0.222 & 0.213 & 0.257 & 0.250 \\
        \midrule
        MUTAG & 0.182 & 0.186 & 0.228 & 0.127 \\
        BZR & 0.182 & 0.192 & 0.198 & 0.128 \\
        COX2 & 0.182 & 0.189 & 0.201 & 0.131 \\
        PROTEINS & 0.185 & 0.202 & 0.215 & 0.161 \\
        \bottomrule
    \end{tabular}

    \label{tab:time_comparison}
\end{table}

\section{Conclusion}
In this paper, we propose MSGCOT, a novel multi-scale prompt chain framework that addresses the limitation of single-granularity prompts in existing graph prompt tuning methods. The proposed low-rank coarsening network captures hierarchical structural features while remaining parameter-efficient. Inspired by the human thought process, MSGCOT progressively introduces information from coarse to fine granularity during prompt generation, leading to more refined prompts. Experimental results demonstrate that MSGCOT consistently outperforms single-granularity approaches and achieves superior performance.

\begin{acks}
This work was supported in part by the National Natural Science Foundation of China under Grants 62425605, 62133012, and 62303366, and in part by the Key Research and Development Program of Shaanxi under Grants 2025CY-YBXM-041, 2022ZDLGY01-10, and 2024CY2-GJHX-15.
\end{acks}
\bibliographystyle{ACM-Reference-Format}
\balance
\bibliography{www2026}

\clearpage
\appendix

\section{Experiment Setting}

\subsection{Dataset Details}

\begin{table}[h]
    \centering
    \caption{Dataset statistics}
\renewcommand{\arraystretch}{1.3} 
\setlength\tabcolsep{1.0pt} 
    \begin{tabular}{lccccccc}
        \toprule
        Datasets & Graphs  & Nodes & Edges & Feats & Classes & Task* (N/G) \\
        \midrule
        Cora & 1  & 2,708 & 5,429 & 1,433 & 7 & N \\
        Citeseer & 1  & 3,327 & 4,732 & 3,703 & 6 & N \\
        Pubmed & 1  & 19,717 & 88,648 & 500 & 3 & N \\
        Photo & 1  & 7,650 & 238,162 & 745 & 8 & N \\
        \midrule
        MUTAG & 188  & 17.9 & 18.9 & 7 & 2 & G \\
        COX2 & 467  & 41.2 & 43.5 & 3 & 2 & G \\
        BZR & 405  & 35.8 & 38.4 & 3 & 2 & G \\
        PROTEINS & 1,113 & 39.1 & 72.8 & 4 & 2 & G \\
        \bottomrule
    \end{tabular}
\end{table}

\subsection{Hyperparameter Setting}
\label{hyperparaset}
As shown in the Table \ref{hyperpara}, we provide detailed parameter configurations for each dataset. cr represents the coarsening rate, cl represents the number of coarsening layers, and r represents the hidden layer dimension of the low-rank coarsening network.

\begin{table}[h]
    \caption{Hyperparameter settings on different datasets.}

    \centering
\renewcommand{\arraystretch}{1.3} 
     \begin{tabular}{lccccccc}
        \toprule
        Datasets & c\_r  & c\_l & r & lr & wd & encoder\_layer \\
        \midrule
        Cora & 0.1  & 2 & 8 & 1e-3 & 0 & 2 \\
        Citeseer & 0.1  & 2 & 8 & 1e-3 &0 & 2 \\
        Pubmed & 0.1  & 1 & 8 & 1e-3 & 0 & 3 \\
        Photo & 0.3  & 2 & 8 & 1e-3 & 0& 2 \\
        \midrule
        MUTAG & 0.05  & 2 & 1 & 1e-3  & 1e-4 & 3 \\
        COX2 & 0.01  & 2& 1 & 1e-3  & 1e-4 & 3 \\
        BZR & 0.01  & 2 & 1 & 1e-3  & 1e-4 & 3 \\
        PROTEINS & 0.1 & 2 & 1 & 1e-3  & 1e-4 & 3 \\
        \bottomrule
    \end{tabular}
    \label{hyperpara}
\end{table}

\section{More Experiments}

\subsection{Effectiveness on different pre-training strategies}
\label{pre+graphcl}

To confirm the generalizability of our method across different pre-training strategies, we employ a GraphCL-pretrained encoder and compare it with several state-of-the-art prompt learning methods, all of which are pre-training-agnostic. As shown in the Table ~\ref{node-class} and Table ~\ref{graph-class}, MSGCOT achieves the best performance on 7 out of 8 benchmark datasets, with particularly significant improvements of +18.23\% and +25.63\% in absolute accuracy on biochemical datasets (COX2 and BZR), respectively. The results indicate that our proposed multi-scale graph chain-of-thought framework does not rely on specific pre-training objectives, validating MSGCOT's potential as a universal prompting framework.

\begin{table}[h]
\caption{Performance comparison on node classification tasks. The best and second-best results are highlighted in bold and underlined, respectively.}
\centering
\renewcommand{\arraystretch}{1.3}
\setlength\tabcolsep{1.5pt}
\small
\begin{tabular}{l|cccc}
    \midrule
 & \textbf{Cora} & \textbf{Citeseer} & \textbf{Pubmed} & \textbf{Photo} \\
    \midrule
GPF+ & 62.47$\pm$14.24 & 49.10$\pm$9.27 & 54.75$\pm$10.56 & 57.99$\pm$10.13 \\
EdgePrompt+ & 63.08$\pm$13.37 & \underline{50.53$\pm$9.15} & 56.17$\pm$9.93 & 57.97$\pm$10.19 \\
GCOT & \underline{63.68$\pm$13.79} & 50.00$\pm$10.20 & \underline{56.42$\pm$10.41} & \textbf{60.41$\pm$10.82} \\
    \midrule
\textbf{MSGCOT} & \textbf{64.24$\pm$17.20} & \textbf{50.56$\pm$9.46} & \textbf{56.69$\pm$12.92} & \underline{58.45$\pm$10.80} \\
    \bottomrule
\end{tabular}
\label{node-class}
\end{table}

\begin{table}[h]
\caption{Performance comparison on graph classification tasks. The best and second-best results are highlighted in bold and underlined, respectively.}
\centering
\renewcommand{\arraystretch}{1.3}
\setlength\tabcolsep{1.5pt}
\small
\begin{tabular}{l|cccc}
    \midrule
 & \textbf{MUTAG} & \textbf{COX2} & \textbf{BZR} & \textbf{PROTEINS} \\
    \midrule
GPF+ & 55.12$\pm$15.79 & 56.35$\pm$16.57 & \underline{49.63$\pm$15.17} & 56.06$\pm$9.49 \\
EdgePrompt+ & 55.60$\pm$14.41 & \underline{56.46$\pm$16.37} & 49.44$\pm$14.91 & \underline{56.33$\pm$9.60} \\
GCOT & \underline{55.75$\pm$14.35} & 52.37$\pm$15.51 & 49.61$\pm$8.48 & 55.12$\pm$10.54 \\
    \midrule
\textbf{MSGCOT} & \textbf{56.42$\pm$15.09} & \textbf{74.58$\pm$4.16} & \textbf{75.26$\pm$4.74} & \textbf{59.02$\pm$6.58} \\
    \bottomrule
\end{tabular}
\label{graph-class}
\end{table}

\begin{figure}[h]
\centering
\includegraphics[width=0.48\textwidth]{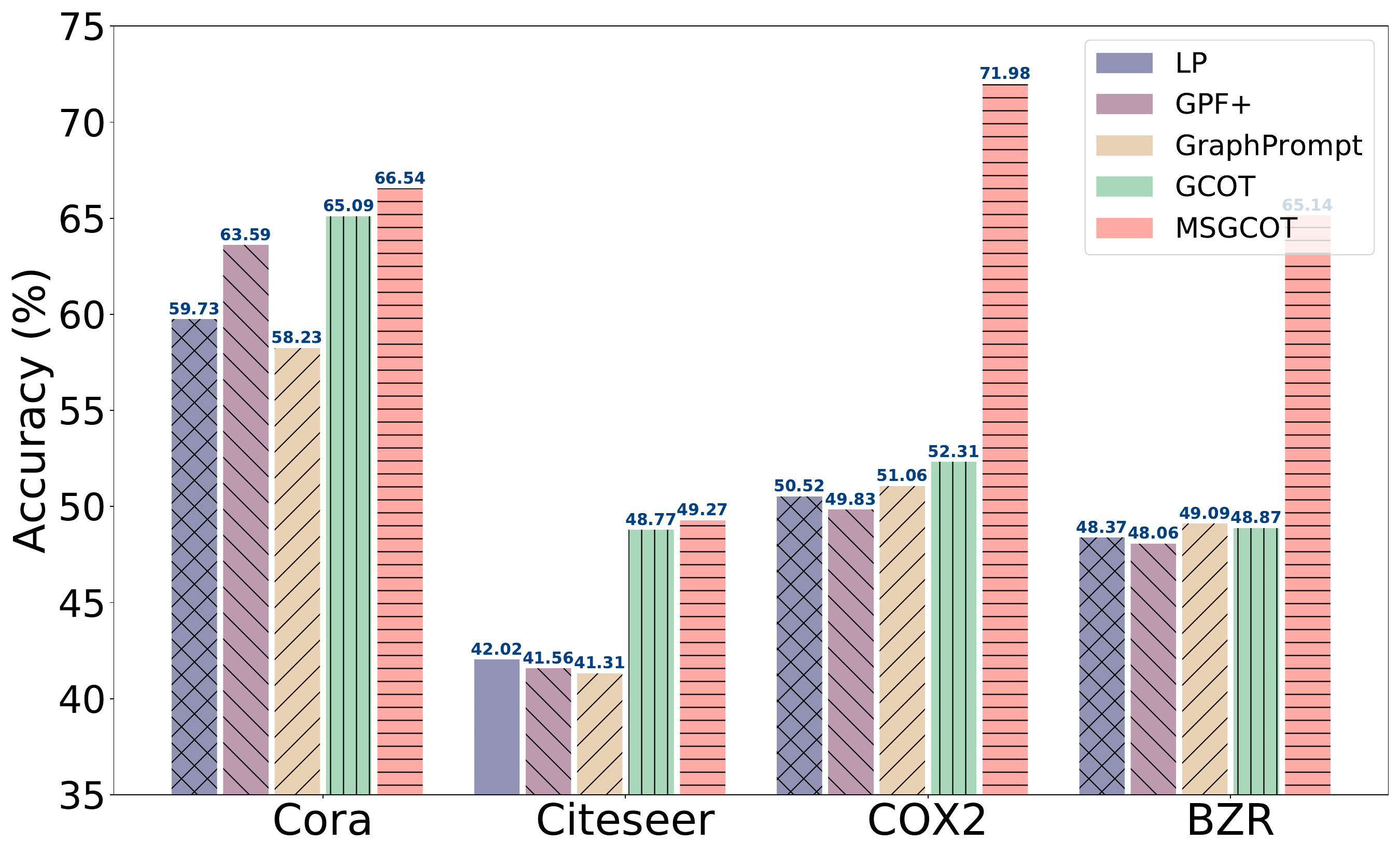}
\caption{The effectiveness of GAT encoders of MSGCOT.}
\label{fig:gat_effect}
\end{figure}

\subsection{Effectiveness on different graph encoders}
To validate the effectiveness of MSGCOT across different encoders, we replaced the original GCN encoder with a graph attention encoder and conducted one-shot node classification experiments on the node classification datasets Cora and Citeseer, alongside the graph classification datasets COX2 and BZR. We present the mean across 100 trials. As shown in Figure \ref{fig:gat_effect}, our approach maintains optimal performance across various datasets under the GAT configuration. This demonstrates that our method constitutes a generalised graph prompt approach whose efficacy remains independent of specific graph encoder architectures.

\subsection{The Performance on Heterophilous Graph Datasets}

\begin{table}[h]
\caption{Performance comparison on heterophilic node classification tasks. The best and second-best results are highlighted in bold and underlined, respectively.}
\centering
\renewcommand{\arraystretch}{1.3}
\setlength\tabcolsep{1.5pt}
\small
\begin{tabular}{l|cccc}
    \midrule
& \textbf{Texas} & \textbf{Cornell} & \textbf{Wisconsin} & \textbf{Chameleon} \\
\midrule
LP & 30.26$\pm$16.19 & 22.62$\pm$6.04 & 24.29$\pm$6.16 & 26.61$\pm$5.89 \\
GraphPrompt & 27.83$\pm$13.92 & 22.38$\pm$6.15 & 23.97$\pm$5.86 & 25.93$\pm$6.21 \\
GPF+ & 26.46$\pm$14.80 & \underline{24.07$\pm$8.96} & 24.12$\pm$7.97 & 26.36$\pm$4.88 \\
GCOT & \underline{29.37$\pm$14.76} & \underline{24.16$\pm$6.88} & \underline{24.00$\pm$6.02} & \underline{27.01$\pm$5.62} \\
\midrule
\textbf{MSGCOT} & \textbf{30.29$\pm$15.42} & \textbf{24.68$\pm$6.82} & \textbf{25.25$\pm$7.39} & \textbf{27.27$\pm$6.29} \\
\bottomrule
\end{tabular}
\label{tab:node-class}
\end{table}
 To evaluate the effectiveness of MSGCOT in heterophilous graph scenarios, we conducted experiments across multiple distinct heterophilous graph datasets, maintaining consistent experimental settings with previous work. Compared to existing methods, we achieved state-of-the-art performance on four datasets—Texas, Wisconsin, Cornell, and Chameleon, validating the efficacy of our multi-scale graph prompt mechanism. These results demonstrate that our multi-scale prompting approach is compatible with both homophilous and heterophilous graphs.

\subsection{Serial vs. Parallel Prompt Fusion Strategies}

We compared the performance of serial (coarse-to-fine) and parallel multi-scale prompt fusion strategies. Experimental results show that the serial strategy consistently outperforms the parallel strategy across multiple datasets, validating the effectiveness of the coarse-to-fine reasoning process.

\begin{table}[h]
\centering
\caption{Performance comparison between serial and parallel prompt fusion strategies}
\begin{tabular}{l|ccc}
\toprule
Strategy & Cora & Citeseer & PROTEINS \\
\midrule
Parallel & $59.85 \pm 15.97$ & $45.07 \pm 10.79$ & $54.52 \pm 8.91$ \\
Serial & $\mathbf{62.13 \pm 17.53}$ & $\mathbf{49.05 \pm 11.41}$ & $\mathbf{57.83 \pm 2.71}$ \\
\bottomrule
\end{tabular}
\end{table}

\subsection{Low-Rank vs. Full-Parameter Coarsening Network}

We compared the performance of the low-rank designed coarsening network with a full-parameter version. The low-rank design significantly outperforms the full-parameter version in parameter efficiency while maintaining competitive performance.

\begin{figure}[ht]
\centering
\includegraphics[width=0.48\textwidth]{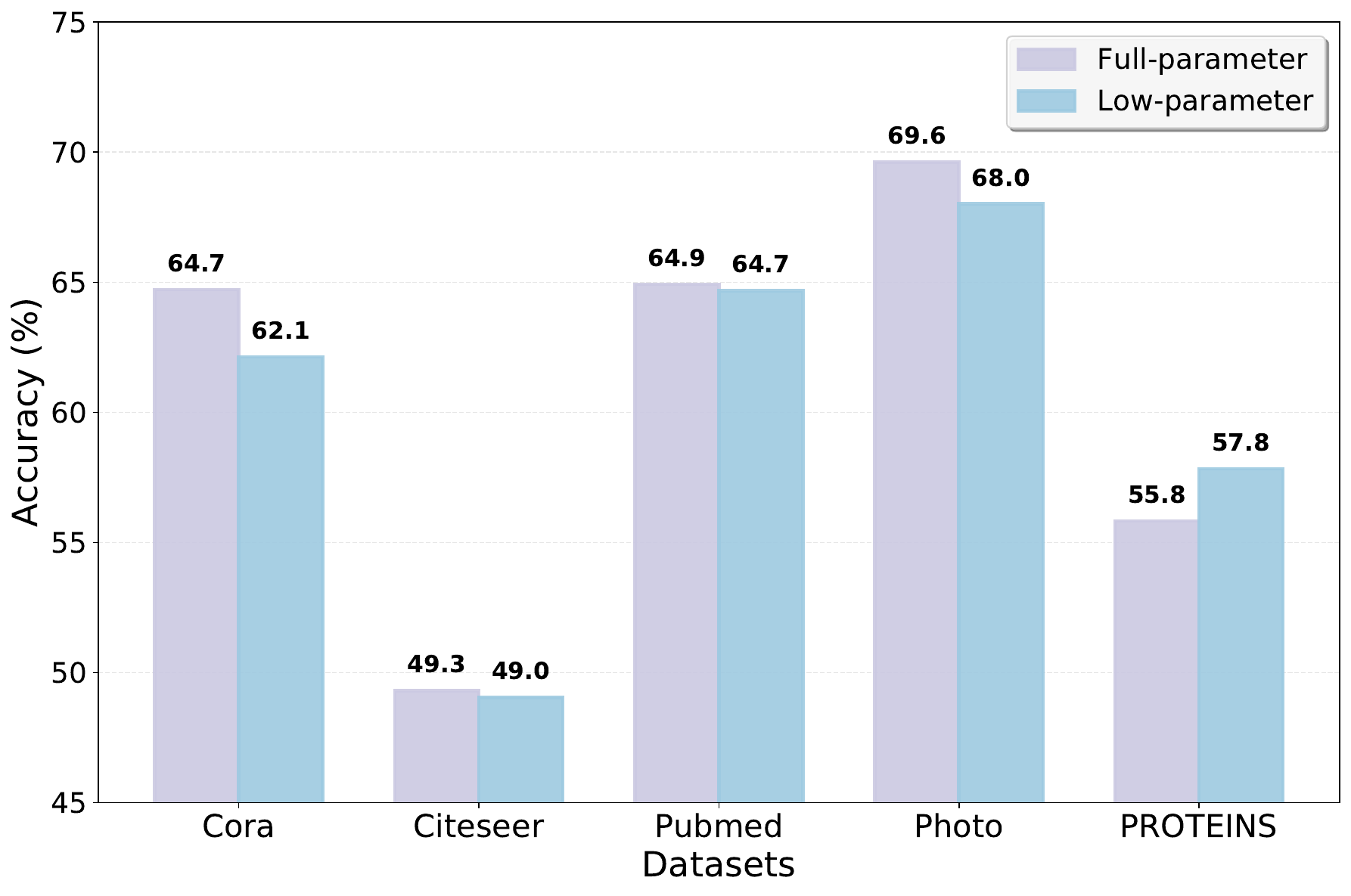}
\caption{Performance comparison between low-rank and full-parameter coarsening networks.}
\label{fig:gat_effect}
\end{figure}

\subsection{Prompt Abalation Experiments}

\para{Prompt Filtering (Top-K).} We tested a prompt selection strategy based on Top-K sampling. Experiments show that selective prompting improves efficiency while maintaining performance.

\begin{table}[h]
\centering
\caption{Performance comparison of Top-K prompt filtering strategies}
\begin{tabular}{l|ccc}
\toprule
Top-K & Cora & Citeseer & PROTEINS \\
\midrule
3 & $59.77 \pm 16.43$ & $48.55 \pm 10.52$ & $54.59 \pm 6.51$ \\
5 & $59.89 \pm 15.66$ & $48.60 \pm 10.61$ & $56.49 \pm 7.34$ \\
10 & $60.20 \pm 15.97$ & $48.34 \pm 10.79$ & $56.33 \pm 7.45$ \\
\bottomrule
\end{tabular}
\end{table}

\para{Multi-Layer Prompt Concatenation.} We concatenate prompts from all layers as the final prompt. Experimental results show that this strategy does not lead to significant performance improvement, indicating that subsequent layers in the chain-of-thought reasoning naturally incorporate information from previous layers.

\begin{table}[h]
\centering
\caption{Performance comparison between multi-layer prompt concatenation and the original method}
\begin{tabular}{l|ccc}
\toprule
Method & Citeseer & Photo & PROTEINS \\
\midrule
Concatenation & $48.54 \pm 9.86$ & $67.99 \pm 10.25$ & $58.43 \pm 16.26$ \\
Original & $\mathbf{49.05 \pm 11.41}$ & $\mathbf{68.01 \pm 10.93}$ & $57.22 \pm 7.03$ \\
\bottomrule
\end{tabular}
\end{table}

\subsection{Large-Scale Dataset Experiment}

We conducted extended experiments on the ogbn-arxiv dataset. MSGCOT demonstrates good scalability.

\begin{table}[h]
\centering
\caption{Performance comparison on the ogbn-arxiv dataset}
\begin{tabular}{l|c}
\toprule
Method & ogbn-arxiv \\
\midrule
GPrompt & $17.74 \pm 3.61$ \\
GCOT & $18.85 \pm 3.66$ \\
MSGCOT & $\mathbf{20.66 \pm 3.90}$ \\
\bottomrule
\end{tabular}
\end{table}

\subsection{Graph Classification on Social Network }

We supplemented graph classification experiments on the IMDB-BINARY and IMDB-MULTI datasets, further validating the effectiveness of the method on diverse graph structures.

\begin{table}[h]
\centering
\caption{Performance comparison on social network graph classification datasets}
\begin{tabular}{l|ccc}
\toprule
Dataset & GPF+ & GPrompt & MSGCOT \\
\midrule
IMDB-BINARY & $50.83 \pm 6.58$ & $52.21 \pm 5.34$ & $\mathbf{54.21 \pm 6.89}$ \\
IMDB-MULTI & $33.65 \pm 5.23$ & $34.59 \pm 2.90$ & $\mathbf{35.30 \pm 4.01}$ \\
\bottomrule
\end{tabular}
\end{table}

\subsection{The Performance on Different Pooling Method}

We compared the performance using DiffPool as the coarsening mechanism. Experiments show that directly using pre-trained representations combined with the low-rank coarsening network avoids overfitting caused by excessive compression, yielding better performance.

\begin{table}[h]
\centering
\caption{Comparison experiment with DiffPool}
\begin{tabular}{l|cc}
\toprule
Method & Cora & MUTAG \\
\midrule
MSGCOT (DiffPool) & $48.83 \pm 12.33$ & $35.31 \pm 7.90$ \\
MSGCOT (Original) & $\mathbf{62.13 \pm 17.53}$ & $\mathbf{63.54 \pm 14.94}$ \\
\bottomrule
\end{tabular}
\end{table}

\end{document}